\documentclass[
aps,twocolumn, superscriptaddress, superscriptreference,
reprint
]{revtex4-2}
\usepackage[utf8]{inputenc}
\usepackage[T1]{fontenc}
\usepackage[]{graphicx}
\usepackage{grffile}
\usepackage{longtable}
\usepackage{wrapfig}
\usepackage{rotating}
\usepackage{amsmath,bm}
\usepackage{textcomp}
\usepackage{amssymb}
\usepackage{capt-of}
\usepackage{soul}
\usepackage{natbib}
\setcitestyle{square}
\usepackage[dvipsnames]{xcolor}
\usepackage[breakable,most,many]{tcolorbox}
\usepackage{ulem}
\usepackage{lineno}
\usepackage{multirow}
\usepackage{amsmath}
\usepackage{amssymb}
\usepackage{array}
\usepackage[colorlinks, linkcolor=blue!50!black, urlcolor=blue!50!black, citecolor=blue!50!black]{hyperref}

\usepackage{pdfpages}
\usepackage{pgffor}
\makeatletter
\AtBeginDocument{\let\LS@rot\@undefined}
\makeatother

\begin{document}

\title{Latent Thermodynamic Flows: Unified Representation Learning and Generative Modeling of Temperature-Dependent Behaviors from Limited Data}

\author{Yunrui Qiu}
\affiliation{Institute for Physical Science and Technology, University of Maryland, College Park, MD, 20742, USA}
\affiliation{Institute for Health Computing, University of Maryland, Bethesda, MD, 20852, USA}

\author{Richard John}
\affiliation{Institute for Physical Science and Technology, University of Maryland, College Park, MD, 20742, USA}
\affiliation{Department of Physics and Institute for Physical Science and Technology, University of Maryland, College Park, MD, 20742, USA}

\author{Lukas Herron}
\affiliation{Institute for Physical Science and Technology, University of Maryland, College Park, MD, 20742, USA}
\affiliation{Institute for Health Computing, University of Maryland, Bethesda, MD, 20852, USA}
\affiliation{Biophysics Program and Institute for Physical Science and Technology, University of Maryland, College Park, MD, 20742, USA}

\author{Pratyush Tiwary}
\thanks{}
\email{ptiwary@umd.edu}
\affiliation{Institute for Physical Science and Technology, University of Maryland, College Park, MD, 20742, USA}
\affiliation{Institute for Health Computing, University of Maryland, Bethesda, MD, 20852, USA}
\affiliation{Department of Chemistry and Biochemistry and Institute for Physical Science and Technology, University of Maryland, College Park, MD, 20742, USA}

\begin{abstract}
Accurate characterization of the equilibrium distributions of complex molecular systems and their dependence on environmental factors such as temperature is essential for understanding thermodynamic properties and transition mechanisms. Projecting these distributions onto meaningful low-dimensional representations enables interpretability and downstream analysis. Recent advances in generative AI, particularly flow models such as Normalizing Flows (NFs), have shown promise in modeling such distributions, but their scope is limited without tailored representation learning. In this work, we introduce \textbf{La}tent \textbf{T}hermodynamic \textbf{F}lows (LaTF), an end-to-end framework that tightly integrates representation learning and generative modeling. LaTF unifies the State Predictive Information Bottleneck (SPIB) with NFs to simultaneously learn low-dimensional latent representations, referred to as Collective Variables (CVs), classify metastable states, and generate equilibrium distributions across temperatures beyond the training data. The two components of representation learning and generative modeling are optimized jointly, ensuring that the learned latent features capture the system’s slow, important degrees of freedom while the generative model accurately reproduces the system’s equilibrium behavior. We demonstrate LaTF’s effectiveness across diverse systems, including a model potential, the Chignolin protein, and cluster of Lennard-Jones particles, with thorough evaluations and benchmarking using multiple metrics and extensive simulations. Finally, we apply LaTF to a RNA tetraloop system, where despite using simulation data from only two temperatures, LaTF reconstructs the temperature-dependent structural ensemble and melting behavior, consistent with experimental and prior extensive computational results.
\end{abstract}

\maketitle

\section{Introduction}
Quantifying microscopic molecular equilibrium distributions and deriving their macroscopic properties is a ubiquitous task across physics, chemistry, materials science, and biology. For example, predicting the populations of protein conformations is critical for elucidating their functional mechanisms and enabling the design of therapeutic molecules\cite{henzler2007dynamic, shaw2010atomic, lindorff2011fast, aranganathan2025modeling, qiu2024non, lewis2024scalable, fan2024accurate}. Similarly, determination of relative stability for material phases over various thermodynamic conditions is essential for constructing phase diagrams and guiding new material designs\cite{vega2008determination, chew2023phase, schebek2024efficient}. Statistical mechanics provides a rigorous framework for computing microscopic equilibrium distributions under given environmental constraints, from which all thermodynamic observables of interest can then be calculated.

Molecular systems of practical interest possess extremely high-dimensional configurational spaces, making it difficult to sample and express their global equilibrium distributions using traditional tools such as Molecular Dynamics (MD) simulations and standard density estimators. These challenges are further exacerbated when studying environment-dependent behaviors, where simulations and estimations must be repeated for each condition. Even when high-dimensional distributions are successfully obtained, they remain difficult to analyze and interpret. Therefore, evaluating the equilibrium distribution on a set of low-dimensional, physically meaningful collective variables (CVs) is a more tractable strategy. This not only enables reliable quantification of thermodynamic properties but also facilitates understanding of the molecular mechanisms underlying complex dynamical processes\cite{torrie1977nonphysical, grubmuller1995predicting, laio2002escaping, henin2010exploring, tiwary2013metadynamics}. Over the past decades, a range of computational techniques have been developed to sketch the free energy surface (FES) on CVs\cite{henin2022enhanced, mehdi2024enhanced, fan2024rid}.  A widely accepted principle for CV selection is that they should capture the slowest transitions between long-lived metastable states. When carefully chosen, such CVs can be used in conjunction with enhanced sampling methods such as umbrella sampling\cite{torrie1977nonphysical} and metadynamics\cite{laio2002escaping} to efficiently reconstruct the unbiased FES.

\begin{figure*}
    \centering
    \includegraphics[
    width=0.86\textwidth
    ]{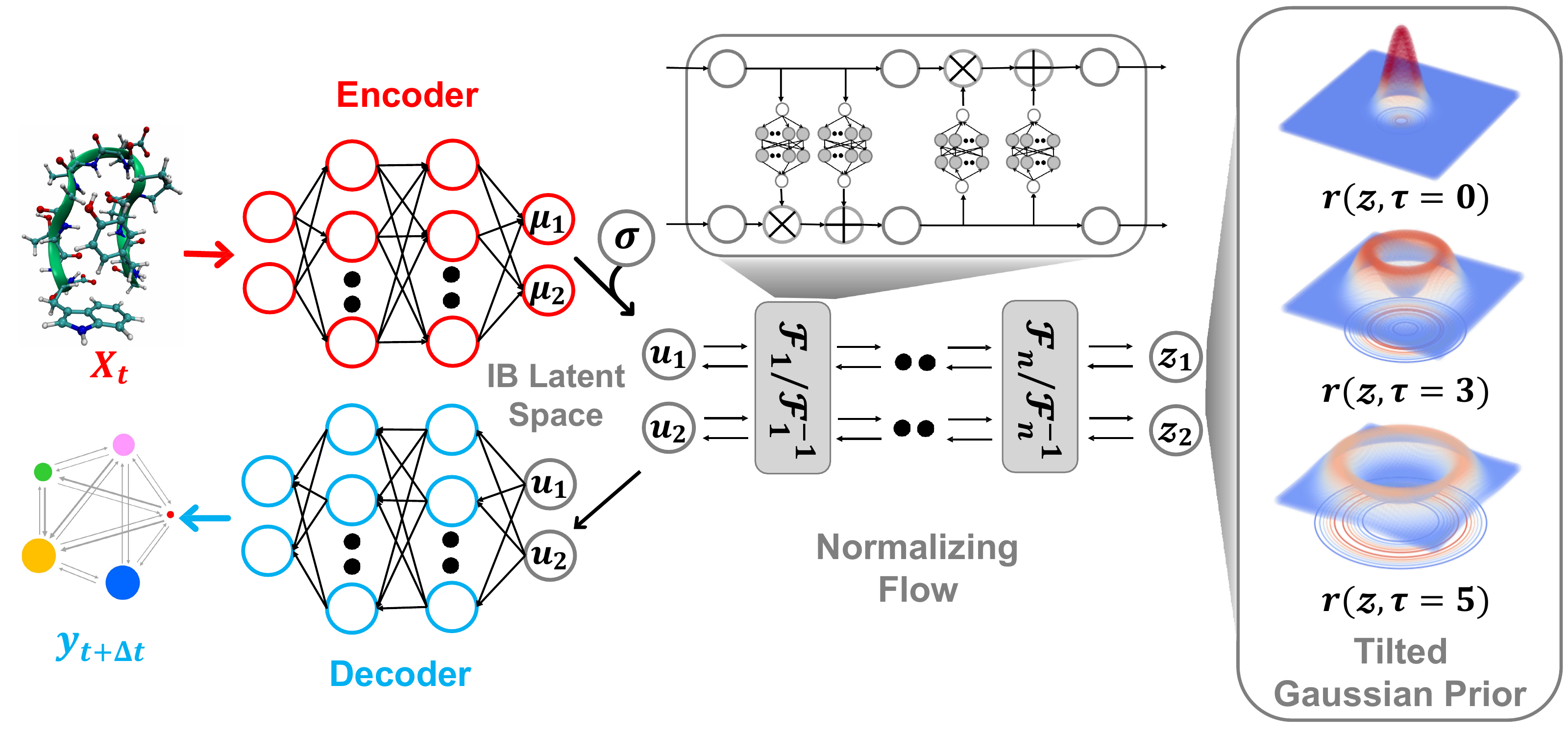}
    \caption{{\bf Architecture of the Latent Thermodynamic Flows (LaTF) model.} The LaTF model consists of three components which are jointly trained together: an encoder that projects molecular descriptors $\bm{X}_{t}$ at time $t$ into a 2D Information Bottleneck latent space $\{\mu_i\}_{i=1}^{2}$; a normalizing flow employing real-valued non-volume preserving transformations $\{\mathcal{F}_{k}/\mathcal{F}
    _{k}^{-1}\}_{k=1}^{n}$ to establish a reversible mapping between the encoded Gaussian distribution $\{u_i\}_{i=1}^{2} \sim \mathcal{N}(u_i; \mu_i, \sigma)$ and the exponentially tilted Gaussian prior $r(\bm{z}, \tau)$; and a decoder that predicts metastable state labels $\bm{y}_{t+\Delta t}$, representing the state into which the input configuration transits after a lag time $\Delta t$. The right most column visualizes the 2D tilted Gaussian prior distributions for several values of the tilting parameter $\tau$, with $\tau=0$ representing the standard Gaussian distribution.}
    \label{fig:LaTF_model}%
\end{figure*}

Recently, the integration of machine learning with physical principles has driven the development of many methods for CV identification and equilibrium distribution prediction. For example, the Information Bottleneck (IB) principle\cite{tishby2000information} and the Variational Approach for Markov Processes\cite{wu2020variational} have inspired numerous methods for CV construction\cite{mardt2018vampnets, wehmeyer2018time, hernandez2018variational, wang2019past, wang2021state}. Several active subfields have emerged from these frameworks, where new methods continue to be developed and applied across chemical, material, and biological systems \cite{lee2024calculating, wang2024local, mehdi2022accelerating, zou2025graph, bonati2021deep, lohr2021kinetic, marques2024covampnet, qiu2025unsupervised, chen2019nonlinear, wang2024information, pengmei2025using, wu2024reaction, teng2025alphafold2, gu2024empowering}.
In parallel, deep generative models have attracted growing interest for generating equilibrium structural ensembles under different environmental conditions. For instance, Boltzmann generators\cite{noe2019boltzmann} and their methodological variants\cite{kohler2020equivariant, dibak2022temperature, klein2024transferable, zheng2024predicting, moqvist2025thermodynamic} have been developed to generate unbiased samples at varying temperatures by constructing invertible flow-based mappings between equilibrium distributions and Gaussian priors. Similarly, Thermodynamic Maps\cite{wang2022data, herron2024inferring} and their extensions\cite{lee2025exponentially, beyerle2024inferring} are capable of inferring how equilibrium distributions vary with thermal variables, including temperature and pressure, using diffusion models trained on limited simulation data. While many generative models struggle with accurately reconstructing all-atom structures, their performance is typically assessed by projecting generated samples onto a few selected CVs\cite{noe2019boltzmann, kohler2020equivariant, dibak2022temperature, zheng2024predicting, klein2024transferable, moqvist2025thermodynamic, wang2022data, herron2024inferring}. However, CV construction and the distribution modeling have so far been treated as two separate tasks and addressed independently.

In the realm of CV construction, most efforts prioritize capturing the slowest dynamical transitions between metastable states, often overlooking their potential for generative tasks. Conversely, generative models emphasize improving sampling fidelity, while giving little attention to the benefits that could arise from learning meaningful CVs, even though they inherently suffer from the curse of dimensionality, where the data required for accurate and generalizable generation grows exponentially with input dimensionality\cite{tsybakov2009nonparametric, biroli2024dynamical, john2025comparison}. This compartmentalized development mirrors similar trends in the broader AI field\cite{vahdat2021score, rombach2022high, su2018f, dao2023flow}. Consequently, uniting representation learning and generative modeling holds both theoretical and practical significance. Projecting data onto a low-dimensional manifold enables the training of more expressive and faithful generative models with much smaller neural network architectures, and accurate modeling of the projected distributions can guide the refinement of these projections.

In this study, we present a unified framework that seamlessly integrates representation learning with generative modeling. Our proposed approach, termed \textit{Latent Thermodynamic Flows} (LaTF), combines the strengths of two powerful models: the \textit{State Predictive Information Bottleneck} (SPIB)\cite{wang2021state} and \textit{Normalizing Flows} (NFs)\cite{papamakarios2021normalizing, dinh2016density} (see model architecture in Fig.~\ref{fig:LaTF_model}). SPIB has demonstrated effectiveness across diverse systems in extracting CVs that capture the slowest dynamical modes from high-dimensional molecular descriptors, while also partitioning configurations into metastable states\cite{gu2024empowering, lee2024calculating, wang2024local, mehdi2022accelerating, zou2025graph}. NFs are powerful generative models that have been applied to approximate complex molecular equilibrium distributions and free energy calculations\cite{noe2019boltzmann, kohler2020equivariant, dibak2022temperature, klein2024transferable, wirnsberger2020targeted, rizzi2021targeted, olehnovics2024assessing}. By employing the NF as a bijective transformation between the latent IB distribution and a prior distribution, we formulate a unified objective that enables the simultaneous training of both SPIB and the NF. We show that the joint training scheme offers benefits complementary to SPIB and NF: it facilitates the optimization of the encoder and decoder, leading to improved delineation of metastable state boundaries, and enables explicit quantification and accurate sampling of the stationary distribution over physically meaningful CVs.

Additionally, in place of the conventional standard Gaussian prior, we employ an exponentially tilted Gaussian distribution for the NF, which expands the volume of high-density regions to facilitate metastable state separation and enable more physically realistic interpolation of transition pathways. By further introducing a temperature-steerable parameter into the tilted prior distribution, LaTF reliably captures nontrivial variations in the IB FES across a broad temperature range, even when trained on data from only a few temperatures. We validate the broad applicability of LaTF across three diverse systems: the Chignolin protein, a cluster of Lennard-Jones particles, and an RNA GCAA tetraloop. In each case, we benchmark our results against extensive computational results or previous experimental studies. Notably, LaTF predicts the temperature-dependent FES and melting temperature of the RNA tetraloop using simulation data collected at only two temperatures, showing agreement with established references. Therefore, we expect that LaTF will serve as a versatile framework for complex systems, providing robust end-to-end capabilities for CVs extraction, metastable state identification, pathway interpolation, equilibrium distribution estimation, and temperature-dependent thermodynamic property prediction.

\section{Results}
\subsection{Setting up Latent Thermodynamic Flows (LaTF): Unifying Representation and Generation with tilted Gaussian Prior}
\label{sec:setup}
A schematic illustration of the LaTF model architecture is shown in Fig.~\ref{fig:LaTF_model}. The LaTF model inherits its ability to identify meaningful CVs and metastable states for complex molecular systems from the well-established SPIB model. SPIB employs a variational autoencoder-like architecture that encodes high-dimensional molecular descriptor $\bm{X}_t$ into a low-dimensional IB space $\bm{z}$ and decodes it to predict the metastable state $\bm{y}_{t+\Delta t}$ into which the input configuration transits after a lag time $\Delta t$. In line with the IB principle\cite{tishby2000information}, SPIB adopts the following loss function:

\begin{small}
\begin{equation}\label{spib_obj}
\begin{split}
    \mathcal{L}_{IB} = -\int d \bm{z}\  p_{\theta}(\bm{z}|\bm{X}_{t}) \Big [\log q_{\theta} (\bm{y}_{t+\Delta t}|\bm{z}) -\beta \log\frac{p_{\theta}(\bm{z}|\bm{X}_{t})}{r(\bm{z})} \Big]
\end{split}
\end{equation}
\end{small}

where $p_{\theta}(\bm{z}|\bm{X}_{t})$ denotes the posterior distribution parameterized by the Gaussian encoder, $q_{\theta} (\bm{y}_{t+\Delta t}|\bm{z})$ represents the decoder-predicted probability of the future states ($\theta$ denotes neural network parameters), and $r(\bm{z})$ is the prior distribution over IB space, defined as a modified variational mixture of posteriors prior (VampPrior). The loss function includes $\beta$, a tunable parameter that balances future-state prediction accuracy with regularization of the encoded posterior towards Gaussian mixture prior. Once trained, the learned IB space ideally captures slow inter-state transitions with timescales longer than $\Delta t$ and enables clear separation of metastable states. SPIB is trained through an iterative and self-consistent scheme in which the short-lived states are merged into long-lived ones and input configurations are relabeled on-the-fly with their most probable future-state labels, thereby maximizing state metastability\cite{wang2024information} (see SI for more SPIB details).

Mapping the FES through enhanced sampling on SPIB-CVs  has proven to be highly informative and broadly applicable across diverse systems, including drug binding and unbinding\cite{lee2024calculating}, biomolecular conformational changes\cite{wang2024information, mehdi2022accelerating, van2023boltzmann}, and crystal polymorph nucleation\cite{meraz2024simulating, wang2024local}. While SPIB employs a multi-modal VampPrior distribution to regularize IB space, we replace that with a more expressive generative model, namely the NF. As shown in Fig.~\ref{fig:LaTF_model}, the NF consists of a sequence of bijective transformations that serve as a change of variables, mapping the encoded distribution to an easily sampled prior distribution. We utilize the real-valued non-volume preserving (RealNVP) transformation\cite{dinh2016density}, which offers an explicitly computable Jacobian determinant (more details in Methods). Inspired by previous studies\cite{vahdat2021score, rombach2022high, su2018f, dao2023flow}, we employ the NF $\mathcal{F}_{\theta}$ to model SPIB's posterior distribution $p_{\theta}(\bm{z}|\bm{X}_{t})$, which can be mathematically expressed as an integral involving the Dirac delta function:

\begin{small}
\begin{align}\label{flow_dirac}
    p_{\theta}(\bm{z}|\bm{X}_{t}) = \int d\bm{u} \ \delta \Big(\bm{z} - \mathcal{F}_{\theta}(\bm{u})\Big) p_{\theta}(\bm{u}|\bm{X}_{t}).
\end{align}
\end{small}

Here, we introduce a new random variable $\bm{u}$, which obeys a Gaussian distribution conditioned on the encoded representation, i.e.,  $p_{\theta}(\bm{u}|\bm{X}_t)=\mathcal{N}(\bm{u}; \bm{\mu}_{\theta}(\bm{X}_t), \sigma_{\theta})$, and is flowed to obtain the posterior distribution. The variance $\sigma_{\theta}$ is treated as a learnable, input-independent parameter.
Unlike SPIB, which directly encodes inputs into Gaussian distributions in a brute-force manner, LaTF adopts a two-step transformation, e.g., encoding followed by flow refinement, allowing for a more expressive representation of the encoded distribution and enabling a more accurate alignment between the posterior and the prior. The unified loss function for LaTF is derived by integrating Eqs. \ref{spib_obj} and \ref{flow_dirac} (see more details in the SI):

\begin{small}
\begin{equation}\label{latf_loss}
\begin{split}
    \mathcal{L}_{LaTF} &= -\int d\bm{u} \ p_{\theta}(\bm{u}|\bm{X}_t) \bigg[\log q_{\theta}\Big(\bm{y}_{t+\Delta t}|\mathcal{F}_{\theta}(\bm{u})\Big) \\&
    +\beta\log r\Big(\mathcal{F}_{\theta}(\bm{u})\Big)  + \beta \log \Big|\det [\frac{\partial \mathcal{F}_{\theta}(\bm{u})}{\partial \bm{u}}]\Big| \bigg] 
\end{split}
\end{equation}
\end{small}

While the first term continues to account for the reconstruction error of future-state prediction, the second and third terms jointly regularize the NF to precisely map the IB latent distribution to the prior. Meanwhile, the incorporation of the NF also also supports the utilization of a more generalized and flexible prior, for which we adopt an exponentially tilted Gaussian distribution\cite{floto2023tilted} (see examples in Fig.~\ref{fig:LaTF_model}):

\begin{small}
\begin{equation}\label{tilted_gaussian}
\begin{split}
    &r(\bm{z}, \tau) = \frac{\exp (\tau ||\bm{z}||)}{Z_{\tau}} \cdot \frac{\exp(-\frac{1}{2}||\bm{z}||^{2})}{(2\pi)^{d_z/2}} \\
    &= \frac{1}{Z_{\tau}(2\pi)^{{d_z/2}}} \cdot \exp\Big(-\frac{1}{2} (||\bm{z}||-\tau)^{2}\Big) \cdot \exp(\frac{1}{2}\tau^{2})
\end{split}
\end{equation}
\end{small}

where $\tau$ is the tilting factor, $d_z$ is IB space dimensionality, and $Z_{\tau}$ denotes the analytically tractable normalization constant (see Methods for details). Different from the standard Gaussian, which is a special case of the tilted Gaussian when $\tau = 0$ and has its mass concentrated at a single point, the tilted Gaussian is radially symmetric and reaches its maximum probability along the ring where $||\bm{z}|| = \tau$. This form provides a much larger volume for the high-density region, facilitating better accommodation and separation of complex encoded data. As we will show later, the tilted Gaussian also offers substantial advantages in predicting temperature-dependence of equilibrium distributions.

So far, two parameters are particularly crucial for the LaTF model: the the IB latent dimension $d_z$ and the tilting factor $\tau$. Prior studies have shown that state-label-driven training of SPIB can embed a large number of states (over ten) within a two-dimensional IB space\cite{wang2024information, van2023boltzmann}, therefore all results in this study are obtained using a two-dimensional IB space. Moreover, we select the optimal tilting factor $\tau$ using the Kullback–Leibler (KL) divergence between the generated and reference distributions. The value of $\tau$ that yields the lowest generation divergence will be adopted (more details for LaTF training are presented in Methods).

\subsection{Benchmarking LaTF for Model Potential and Chignolin in Explicit Water}
\label{sec:benchmark}

\begin{figure*}
    \centering
    \includegraphics[
    width=0.9\textwidth
    ]{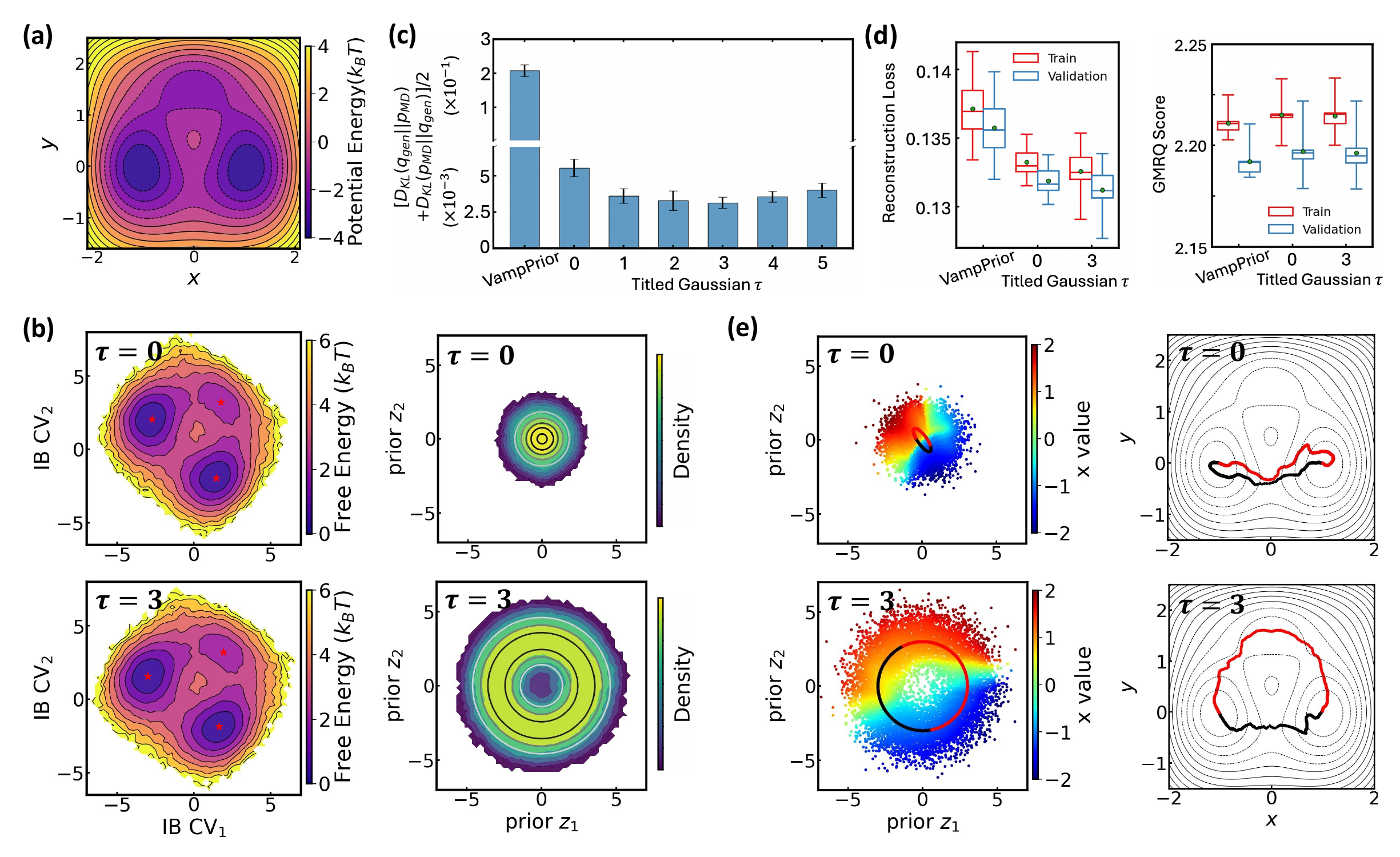}
    \caption{{\bf Benchmarking LaTF performance on 2D three-hole potential system at temperature $\bm{k_BT=1}$.} (a) Analytical potential energy surface. (b) Projections of simulation data onto the IB latent spaces (left) and prior spaces (right) by LaTF models trained with different tilting factors. Asterisks indicate the configurations with the highest likelihood quantified by LaTF in each state. Contour lines of the analytical prior illustrate LaTF’s ability to learn an effective mapping between IB latent and prior distributions. (c) Symmetric KL divergence between the reference IB distribution (from the validation dataset) and the distributions generated by vanilla SPIB with VampPrior and LaTF models with different tilting factors. (d) Comparison of the quality of metastable state assignment between vanilla SPIB and LaTF. All uncertainties in (c) and (d) are derived via five-fold cross-validation. (e) Interpolation of transition pathways in the prior spaces (left), where background shows projected data colored by their $x$-coordinate values. Interpolated samples are mapped back to $x-y$ coordinate space via nearest neighbors in the IB space (right).} 
    \label{fig:three_hole_potential}%
\end{figure*}

\begin{figure*}
    \centering
    \includegraphics[
    width=0.92\textwidth
    ]{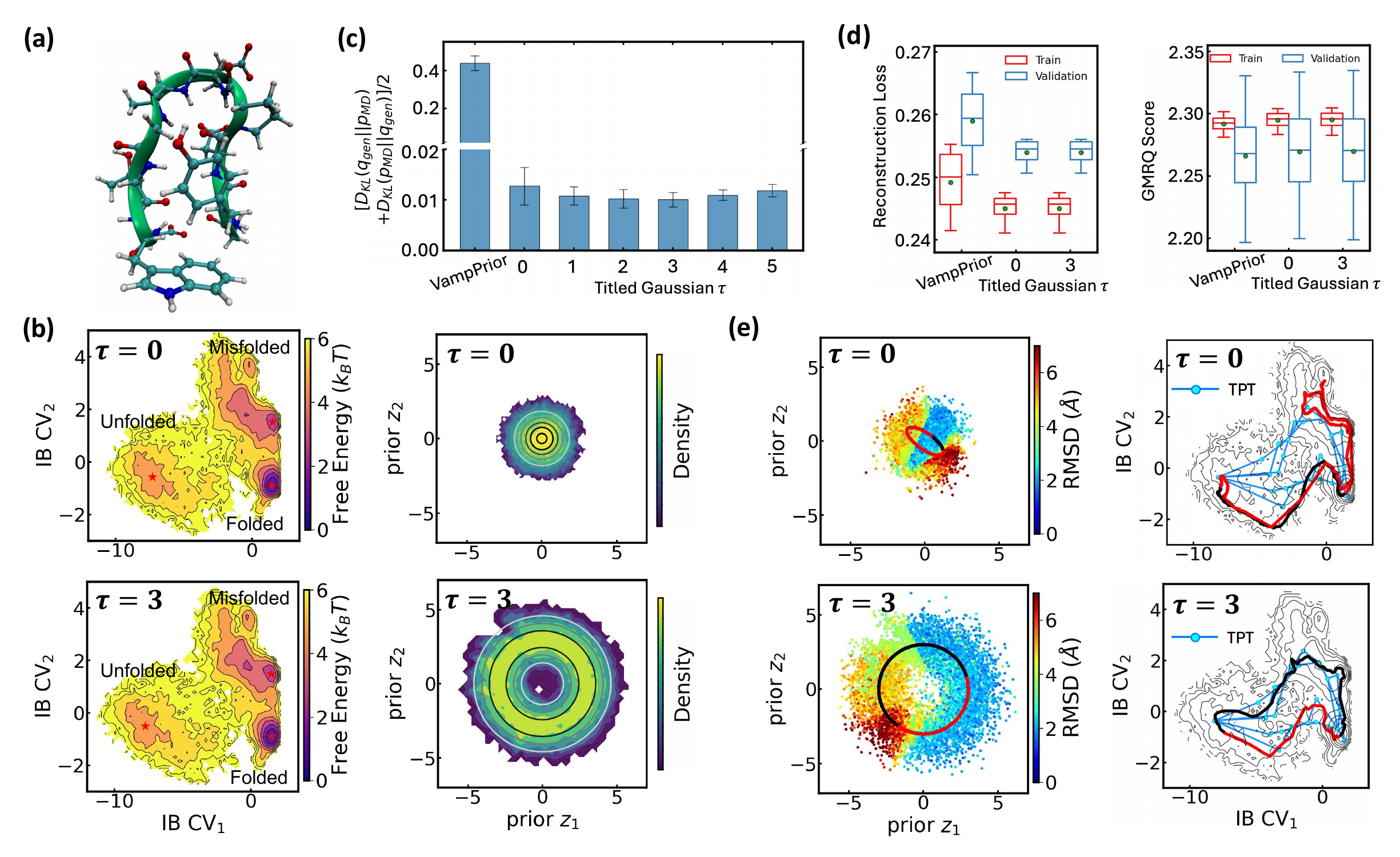}
    \caption{{\bf Benchmarking LaTF performance on Chignolin protein system at temperature 340 K.} (a) PDB structure of the Chignolin protein (PDB: 1UAO). (b) Projections of simulation data onto the IB latent space (left) and prior space (right) by LaTF models trained with different tilting factors. Asterisks indicate the configurations with the highest likelihood quantified by LaTF in each state. Contour lines of the analytical prior illustrate LaTF’s ability to learn an effective mapping between latent and prior distributions. (c) Symmetric KL divergence between the reference IB distribution (from the validation dataset) and the distributions generated by vanilla SPIB with VampPrior and LaTF with different tilting factors. (d) Comparison of the quality of metastable state assignment between vanilla SPIB and LaTF. All uncertainties in (c) and (d) are derived via five-fold cross-validation. (e) Interpolation of transition pathways in the prior space (left), where the background shows projected data colored by the heavy-atom RMSD relative to the PDB structure. The interpolated pathways are flowed to the IB space and compared with the top five highest-flux pathways identified by Transition Path Theory (right). }
    \label{fig:chignolin_340k}%
\end{figure*}


The effectiveness of LaTF and the exponentially tilted Gaussian prior is demonstrated on both a model potential and the well-studied Chignolin protein system, as we have accurate results to compare with. The 2D potential shown in Fig.~\ref{fig:three_hole_potential}(a) features two deep basins connected by two reaction channels, with the upper channel containing a local minimum. Langevin dynamics simulation of a single particle on this potential surface is performed and used to train LaTF models with varying tilting factors (see Methods for simulation and training details). 

Visual inspection (Fig.~\ref{fig:three_hole_potential}(b) and S1) reveals that, after appropriate rotation, the IB spaces learned by LaTF closely resembles the original potential surface, and the flow-transformed distribution in prior space aligns well with the analytical prior, indicating that LaTF correctly learns meaningful IB space and meanwhile accurately captures its distribution. To quantitatively evaluate the performance, we apply three metrics and compare LaTF with a vanilla SPIB model (with VampPrior) trained under identical settings, except without the NF module. The KL divergence between IB distributions derived from generation and simulation data (Fig.~\ref{fig:three_hole_potential}(c)) confirms that the NF model largely improves alignment between the encoded distribution and the prior, with the tilted Gaussian prior outperforming the standard Gaussian. Incorporating the NF and tilted Gaussian also reduces LaTF’s reconstruction loss for future-state prediction (i.e., the first term in Eq.~\ref{latf_loss}) compared to the vanilla SPIB, suggesting improved modeling of inter-state transitions and better metastable state classification (Fig.~\ref{fig:three_hole_potential}(d), left). This is further supported by the improved performance of the resulting Markov State Model (MSM)\cite{prinz2011markov} constructed using state labels derived from the decoder, as assessed by the generalized matrix Rayleigh quotient (GMRQ) score\cite{mcgibbon2015variational} (Fig.~\ref{fig:three_hole_potential}(d), right). The GMRQ quantifies state metastability and the model’s capacity to capture slow dynamics, with higher scores indicating better performance (see Methods for details).
These results demonstrate that the unified training framework promotes mutual enhancement between representation learning and generative modeling, with the encoder, decoder, and NF components performing better when trained together. 

Additionally, the flexibility of the tilted Gaussian enables richer structure in the prior space, which in turn supports more physically realistic interpolations in the IB space. Since the $x$-coordinate is a good proxy for the committor function in this system,  we color the data flowed to the prior space by their $x$-values (Fig.~\ref{fig:three_hole_potential} (e), left column).  The standard Gaussian, which is the most common prior used in generative models to approximate molecular equilibrium distribution, collapses the data into a narrow high-density region, whereas the tilted Gaussian distributes them more uniformly, clearly distinguishing the two reaction channels.  Using the trained NF model to evaluate configuration likelihoods, we identify the most probable states in each basin and interpolate pathways between them. For the standard Gaussian prior we use spherical linear interpolation\cite{shoemake1985animating}, while for the tilted Gaussian prior, taking advantage of its structured form, we directly apply simplest linear interpolation between the central angles and radii of the path endpoints (see SI for details). The standard Gaussian prior produces pathways confined to a single reaction channel, whereas the tilted Gaussian recovers transition pathways across two distinct channels (Fig.~\ref{fig:three_hole_potential}(e), right column).

Extending the above analysis, we further demonstrate LaTF's robustness on a ten-residue protein system, Chignolin (PDB: 1UAO; Fig.~\ref{fig:chignolin_340k}(a))\cite{honda200410}, using a long unbiased MD trajectory ($\sim 35 \mu s$) simulated at 340 K in all-atom resolution including explicit water (see Methods for simulation and LaTF training details). Consistent with earlier results, the LaTF model clearly distinguishes Chingolin's folded, unfolded, and misfolded states in the IB latent space (Fig.~\ref{fig:chignolin_340k}(b)). The close agreement between the generated and reference FES in the IB space (Fig.S2) underscores LaTF’s advantage in optimally leveraging data to accurately model distributions over physically meaningful variables. Meanwhile benefiting from the NF model, which flexibly refines the mismatch between posterior and prior, LaTF outperforms vanilla SPIB in IB distribution approximation, future-state prediction, and metastable state classification (Fig.~\ref{fig:chignolin_340k}(c-d)). 

Notably, the advantages of the tilted Gaussian prior are further emphasized when performing pathway interpolation for the Chignolin protein. As shown in Fig.~\ref{fig:chignolin_340k}(e), visualizing the distribution of heavy-atom root-mean-square deviation (RMSD) relative to the PDB structure in prior space reveals the limitation of the standard Gaussian: it places folded structures in a high-density region, while misfolded and unfolded structures lie at opposite low-density boundaries, connected only by a narrow region that, although critical for folding transitions, is poorly represented. In contrast, the tilted Gaussian spreads the configurations more uniformly, clearly resolving folded and unfolded states and connecting them through two distinct and well-represented transition regions. Even with spherical linear interpolation, the standard Gaussian prior fails to produce pathways consistent with those identified by Transition Path Theory\cite{vanden2006transition, noe2009constructing, qiu2023efficient} (TPT; see SI for more information), whereas simple linear interpolation on the tilted prior yields pathways that align with the TPT results. Further back-mapping of samples interpolated with the tilted prior, using nearest neighbors from simulation data (based on Euclidean distance in IB space), reveals two distinct folding mechanisms that differ in the sequence of H-bond breaking involving residue Asp3, Thr6, and Thr8 (see Fig.S3), consistent with findings from previous studies\cite{trizio2025everything, kang2024computing}. 

Results from both systems confirm the effectiveness of LaTF, which we attribute to its conjugate training framework and structured prior. Many existing approaches employ NFs for full-coordinate structure generation, however, they are generally restricted to implicit-solvent systems, demand extensive preprocessing of molecular Cartesian coordinates and exhibit limited performance in data-sparse yet critical transition regions\cite{noe2019boltzmann, dibak2022temperature, klein2024transferable}. Differently, LaTF serves as a post-analysis framework that efficiently use explicit-solvent simulation data to model equilibrium distributions over a representative latent space, which reduces generative uncertainty and provides an effective distance metric for backmapping to full-coordinate structures. In the next section, we further evaluate LaTF's data efficiency and physical transferability across temperatures.

\subsection{Quantifying LaTF's Ability to Enable Sampling at Out-of-distribution Temperatures}
\label{sec_LaTF_temp_quantify}

\begin{figure*}
    \centering
    \includegraphics[
    width=1.0\textwidth
    ]{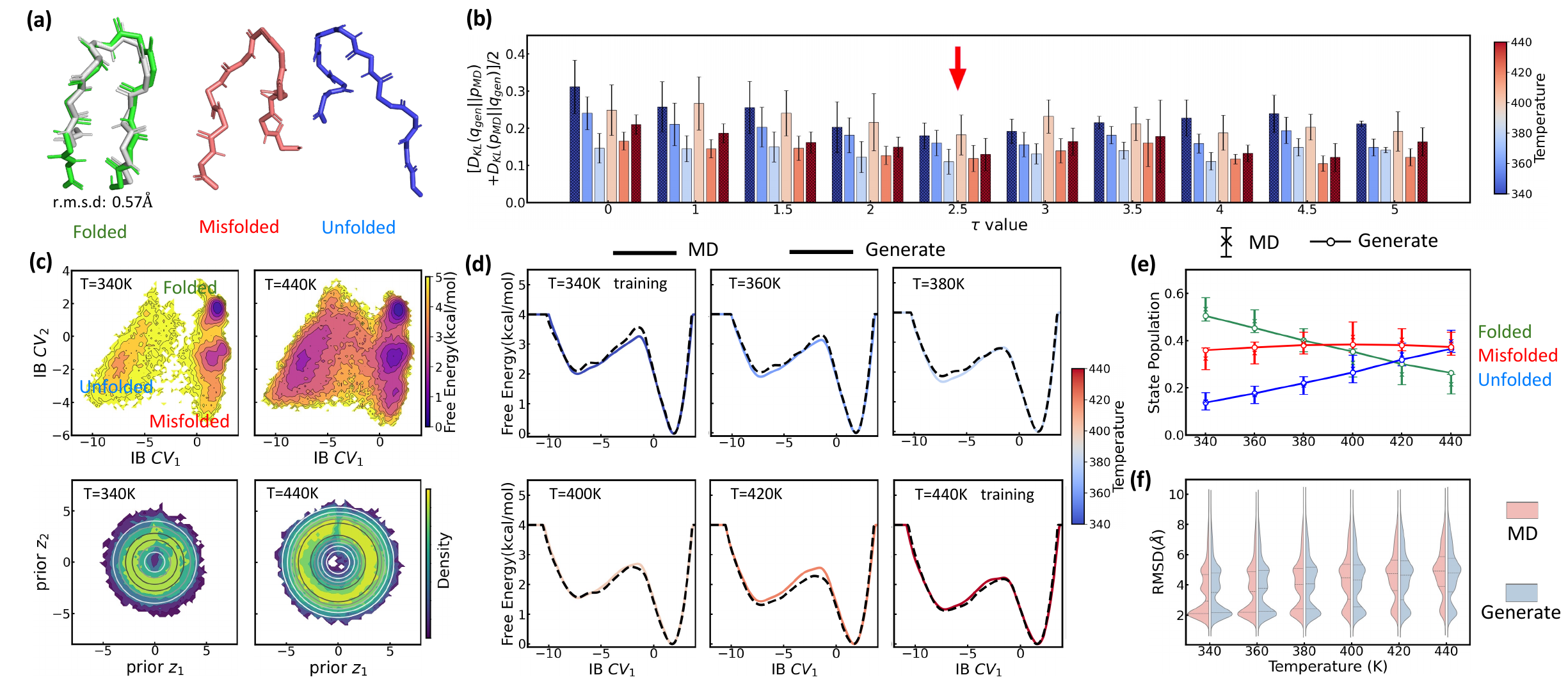}
    \caption{{\bf Evaluating LaTF performance in inferring temperature-dependent equilibrium distributions of Chignolin protein system using MD data at 340 K and 440 K.} (a) Backbone structures of Chignolin with the highest likelihood in three metastable states (colored), shown alongside the PDB structure (gray) for comparison. (b) Symmetric KL divergence between LaTF-generated and simulation-derived IB distributions. For training temperatures (340 K \& 440 K, shaded), divergence is computed on validation data; for other temperatures, it is quantified using the full MD data. Uncertainties are estimated from five-fold cross-validation. The optimal $\tau=2.5$ is chosen (red arrow) to minimize the divergence for only training temperatures. (c) FES constructed from encoded MD data in IB latent space, and density map of flow-transformed encoded data in the prior space for 340 K and 440 K. Analytical priors are shown in contour lines. (d) Comparison of FES along IB CV$_1$ from generated samples (solid) and long MD simulations (dashed) across six temperatures. (e) Populations of metastable states decoded from LaTF-generated samples across six temperatures; reference values and uncertainties are estimated via Bayesian MSMs \cite{metzner2009estimating}. (f) Heavy-atom RMSD distributions relative to PDB structure for MD data and LaTF-generated samples, reconstructed to all-atom structures using nearest neighbors in latent space. Distribution quartiles are marked with dashed lines.} 
    \label{fig:chignolin_340-440k}%
\end{figure*}

\begin{figure*}
    \centering
    \includegraphics[
    width=1.0\textwidth
    ]{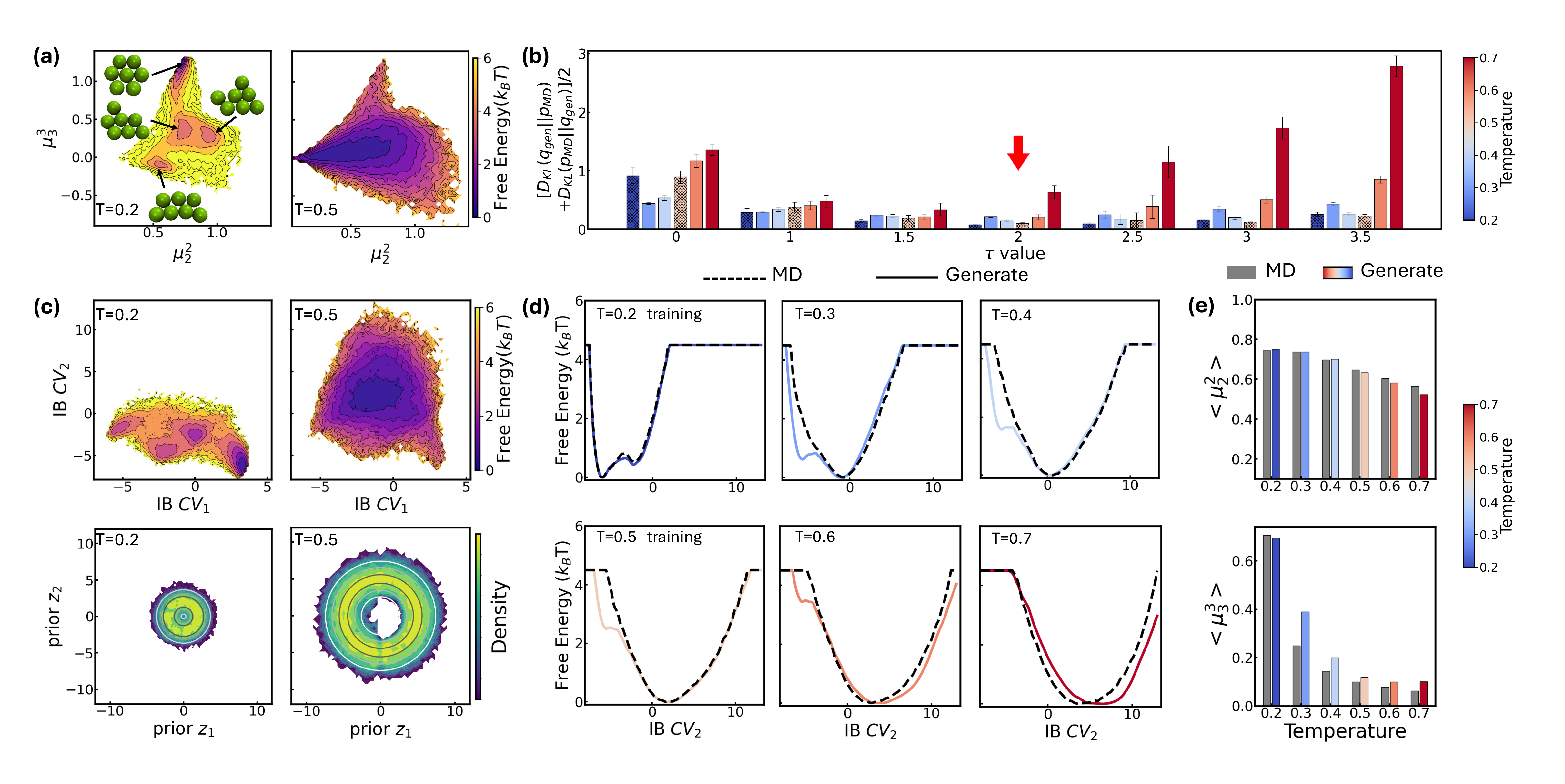}
    \caption{{\bf Inferring the temperature dependence of LJ7 system from data at $0.2\epsilon/k_{B}$ and $0.5\epsilon/k_{B}$ via LaTF.} (a) Projected FES onto the second and third moments of coordination numbers, $\mu_2^2$ and $\mu_3^3$, at $0.2\epsilon/k_B$ and $0.5\epsilon/k_B$. For each metastable state at $0.2\epsilon/k_B$, the structure with the highest likelihood scored by LaTF is shown. (b) Symmetric KL divergence between distributions generated by LaTF and those from MD simulations. For training temperatures ($0.2\epsilon/k_B$ and $0.5\epsilon/k_B$), the divergence is computed using the validation dataset; for other temperatures, the full MD data are used. The optimal $\tau=2$ is chosen (red arrow) to minimize the divergence for training temperatures (temperatures apart from $0.2\epsilon/k_B$ and $0.5\epsilon/k_B$ were not used for optimal $\tau$ calculation). Hatched bars indicated training temperatures. (c) FES in the IB space and density in the prior space estimated from encoded and flow-transformed MD data. Contour lines indicate the analytical tilted prior. (d) Comparison of LaTF-generated FES (solid) with MD-derived FES (dotted) across temperatures. (e) Generative (colored) vs. reference (gray) means of the second and third moments of coordination numbers across temperatures. Generated samples are mapped to full-coordinate structures via nearest neighbors in the IB space.}  
    \label{fig:LJ7_2-5}%
\end{figure*}

\textbf{Temperature-steerable tilted Gaussian prior.} Temperature is a fundamental thermodynamic variable that sets the scale of thermal fluctuations relative to the potential energy. Although the effect of temperature is conceptually simple, when considered in microscopic coordinates and invoking equipartition theorem, its effect on macroscopic thermal properties is highly non-trivial. As a result, the development of methods for predicting how thermodynamic properties respond to changes in temperature has become increasingly important in recent years. Existing generative models typically handle temperature-dependent inference by encoding temperature into the prior distribution (e.g., via variance modulation of a Gaussian prior)\cite{noe2019boltzmann, dibak2022temperature}, or incorporating it explicitly as a model input\cite{janson2025deep}, or employing a combination of both approaches\cite{herron2024inferring, dibak2022temperature}. Here, we adopt the first strategy by introducing a temperature-steerable parameter into LaTF's tilted Gaussian prior to allow inference of temperature-dependent FES in IB space. Specifically, we define the temperature-steerable tilted Gaussian as:

\begin{small}
\begin{equation}\label{tilted_gaussian_temp}
\begin{split}
    &r_{T}(\bm{z}, \tau) = \frac{\exp (\tau ||\bm{z}||)}{Z_{\tau, T}} \cdot \frac{\exp(-\frac{1}{2T}||\bm{z}||^{2})}{(2\pi T)^{d_{z}/2}} \\
    &= \frac{1}{Z_{\tau, T}\ (2\pi T)^{\frac{d_{z}}{2}}}\cdot \exp(-\frac{1}{2T}(||\bm{z}||-T\tau)^{2}\cdot\exp(\frac{1}{2}\tau^{2}T^{2})
\end{split}
\end{equation}
\end{small}

where $T$ is the temperature-steerable parameter and $Z_{\tau, T}$ denotes normalization factor (more details in Methods). Completing the square reveals that both the variance and the radius of maximum probability of $r_{T}(\bm{z}, \tau)$ increase with temperature. This aligns with the physical intuition that higher temperatures amplify entropic effects, potentially shifting and reshaping free energy basins relative to those defined purely by potential energy minima. The proposed formulation thus enables accurate FES modeling while capturing temperature-specific variations. When $\tau=0$, the distribution reduces to a standard Gaussian with variance linearly dependent on $T$, a form adopted in previous generative models\cite{noe2019boltzmann, dibak2022temperature}.

\textbf{Inferring temperature-dependent behaviors of Chignolin protein.} For the first benchmark study, we evaluate LaTF's performance on \textit{three} tasks using the Chignolin protein system (Fig.~\ref{fig:chignolin_340-440k}(a)), comparing against extensive unbiased MD simulations conducted at six temperatures from 340 K to 440 K (see Methods for details). The first task for this benchmark system involves generating FES at intermediate temperatures using long simulation data from only the extremal temperatures, 340 K and 440 K. As before, the optimal $\tau$ is selected as 2.5 to minimize the KL divergence between generated and simulated distributions at the training temperatures (Fig.~\ref{fig:chignolin_340-440k}(b)). Visualizations of the FES in IB space and the associated density in prior space confirm that LaTF simultaneously identifies meaningful CVs and accurately captures the equilibrium distribution (Fig.~\ref{fig:chignolin_340-440k}(c)). The most likely structure in the folded state scored by LaTF aligns closely with the reference NMR structure (Fig.~\ref{fig:chignolin_340-440k}(a)). Notably, the temperature-steerable tilted Gaussian prior ($\tau > 0$) consistently yields lower generation errors than the standard temperature-dependent Gaussian ($\tau = 0$), with similar accuracies observed at both training and unseen temperatures (Fig.~\ref{fig:chignolin_340-440k}(b) and (d)). These results highlight LaTF’s strong capability for generalizable ensemble generation at reduced cost. Meanwhile, LaTF’s decoder can assign generated samples in IB space to metastable states, enabling inference of temperature-dependent state populations. These predictions agree well with those from MSMs built on long MD trajectories using the same state definitions (Fig.~\ref{fig:chignolin_340-440k}(e)). Finally, we reconstruct structural ensembles across temperatures by backmapping generated samples to their nearest neighbors in IB space encoded from training data (based on Euclidean distance). The resulting RMSD distributions closely match those from simulations (Fig.~\ref{fig:chignolin_340-440k}(f)), demonstrating LaTF’s robust generative power of structural ensembles, which stems from both expressive CVs and a powerful latent-to-prior mapping.

The second task for the Chignolin system evaluates LaTF’s performance under limited data availability. We train the LaTF model using only first $1\ \mu$s simulation segments from 340 K and 440 K, where the energy landscape is more thoroughly sampled at 440 K but the unfolded state is sparsely sampled at 340 K. As shown in Fig.S4, incorporating data from both temperatures into training significantly refine the estimation of the low-temperature FES, particularly in the transition and unfolded regions. Remarkably, despite the limited data, LaTF remains robust in extracting meaningful CVs and capturing both the FES topology and its temperature dependence.

We then proceed to the third task, in which LaTF is trained using long simulation data from two relatively high temperatures (e.g., 380 K and 440 K) and used to generate FES at lower temperatures. As shown in Fig. S5, the results are consistent with previous findings and confirm that LaTF captures the correct temperature dependence, although the generation errors increase slightly at lower temperatures. Altogether, these three tasks demonstrate that LaTF with a tilted prior achieves strong physical transferability across temperatures, even under data-scarce conditions.

\textbf{Predicting temperature-driven transitions in the Lennard-Jones 7 Cluster.} LaTF is further evaluated on a multi-body system, the Lennard-Jones 7 (LJ7) cluster, where seven particles interact via the Lennard-Jones potential (see Methods for simulation details)\cite{tribello2010self, schwerdtfeger2024100}. Even though this has fewer atoms than Chigolin, the FES of LJ7 exhibits richer temperature dependence due to competing energetic and entropic contributions. Using physical order parameters (OPs), such as the second and third moments of coordination numbers (details in SI), we observe clear thermally driven transitions and metastable-state flipping (Fig.~\ref{fig:LJ7_2-5}(a)), highlighting strong entropic effect at high temperature and making LJ7 a valuable test case for LaTF’s generation capacity. Trained on data from $0.2\epsilon/k_{B}$ and $0.5\epsilon/k_{B}$, LaTF identifies four metastable states at low temperature and captures their temperature-dependent shifts in the IB space, consistent with the physical OPs (Fig.~\ref{fig:LJ7_2-5}(c)).  Generation KL divergence demonstrates that tilted Gaussians with an appropriate $\tau$ parameter significantly outperform standard Gaussians in predicting FES across temperatures from $0.2\epsilon/k_{B}$ to $0.7\epsilon/k_{B}$ (Fig.~\ref{fig:LJ7_2-5}(b)). This advantage arises from their ability to allow both overlap and separation between priors at different temperatures, while standard Gaussians tend to let high-temperature priors fully cover those at lower temperatures, thereby obscuring entropic distinctions. By selecting $\tau$ based solely on generation error at training temperatures, LaTF generalizes well to unseen temperatures with comparable accuracy, enhancing data efficiency. Notably, LaTF captures the temperature-dependent flipping and shifting of FES basins (Fig.~\ref{fig:LJ7_2-5}(d)), though the most populated metastable state corresponding to hexagonal structures is gradually diminished as temperature rises, whereas MD simulations show it should be sharply suppressed. This limitation may be addressed by refining the choice of $\tau$. Furthermore, structural ensembles reconstructed from generated IB samples via nearest-neighbor matching in the training data yield coordination number moments consistent with MD simulations (Fig.~\ref{fig:LJ7_2-5}(e)). Similar results are obtained when training LaTF with data at $0.2\epsilon/k_{B}$ and $0.7\epsilon/k_{B}$ (Fig.S6). All above results validate that LaTF exhibits strong transferability, precisely inferring equilibrium distributions at unseen temperatures and significantly improving data efficiency.

\subsection{Exploring Temperature-Dependent RNA Free Energy Landscapes with LaTF}
\label{sec_LaTF_RNA}

\begin{figure*}
    \centering
    \includegraphics[
    width=0.8\textwidth
    ]{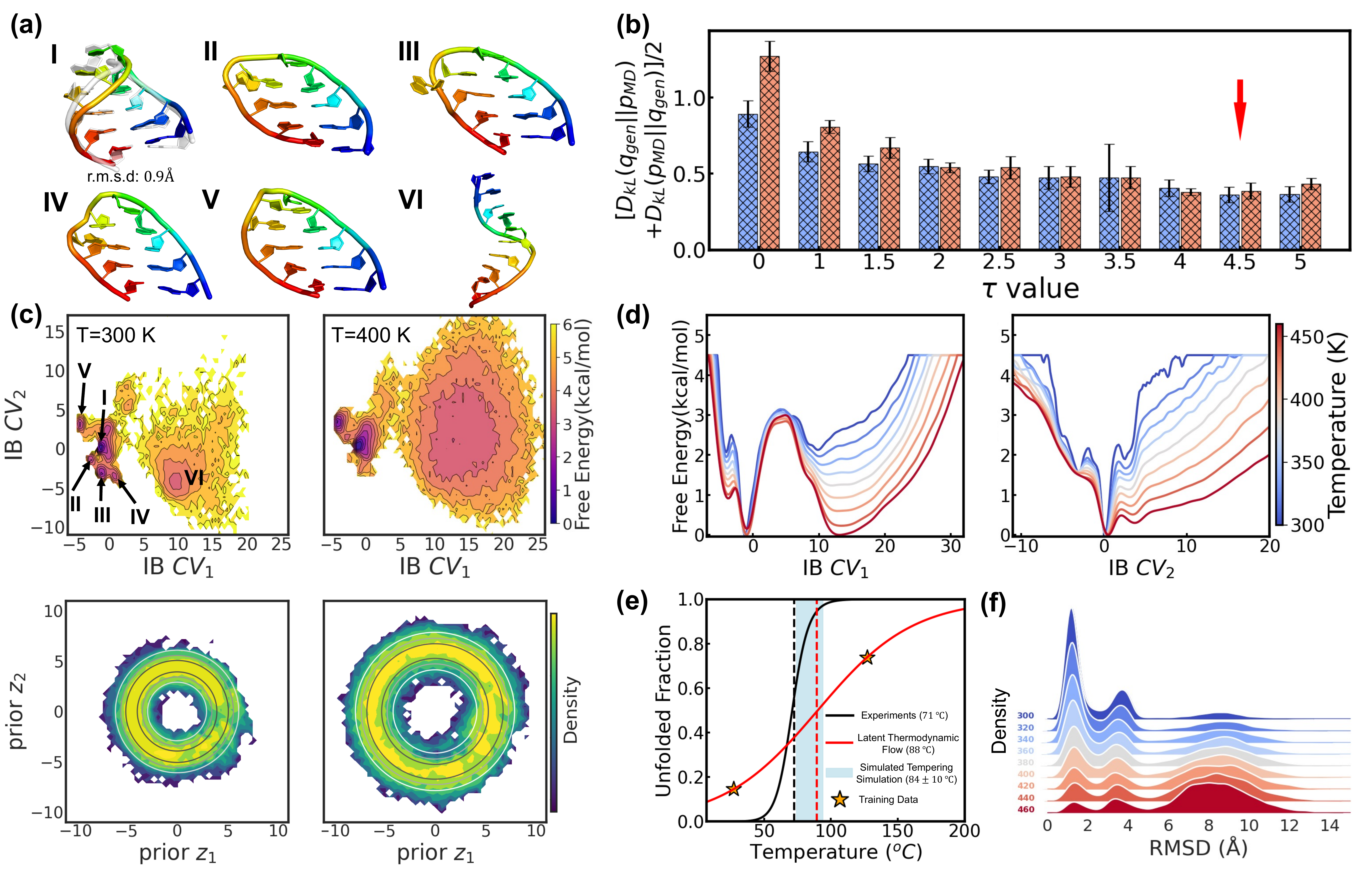}
    \caption{{\bf Inferring the temperature dependence of GCAA RNA tetraloop system from data at 300 K and 400 K via LaTF.} (a) Structure with the highest likelihood scored by LaTF from six metastable states. The NMR structure (gray, transparent) is overlaid with the identified folded structure for comparison. The structures are colored from blue to red from the $5^\prime$ to $3^\prime$ end. (b) Symmetric KL divergence between LaTF-generated IB distributions and those encoded from validation data. The optimal $\tau=4.5$ (indicated by the red arrow) minimizes the KL divergence. (c) FES in the IB space and associate density in the prior space, estimated from encoded and flow-transformed simulation data. Contours represent the analytical tilted prior. (d) Temperature-dependent FES inferred by LaTF in the IB space across a range of 300 K to 460 K at 20 K intervals. (e) Unfolded fraction over temperatures predicted by LaTF (red), compared with the experimental observations (black)\cite{sheehy2010thermodynamic} and simulated tempering results (light blue)\cite{tan2018rna}. Structural ensembles generated by LaTF are reconstructed using nearest neighbors in IB space from simulation data. A conformation is classified as folded if all three base pairs are formed and its all-atom RMSD to any NMR structure is below 4.0~\AA; otherwise, it is considered unfolded. (f) RMSD distributions of LaTF-generated structures relative to the NMR structure across different temperatures.} 
    \label{fig:cgaa_tetraloop_300-400}%
\end{figure*}

We now test LaTF for a significantly more challenging task, i.e.,  simultaneously identifying dynamically meaningful low-dimensional representations and predicting the temperature response for an RNA system. Unlike proteins, which typically fold into a dominant native state within a funnel-shaped energy landscape\cite{frauenfelder1991energy, onuchic1997theory}, RNAs adopt a highly heterogeneous ensemble of structures with multiple metastable states sharing comparable stability, leading to an intrinsically rugged and disordered energy landscape\cite{chen2000rna, kuhrova2016computer, pagnani2000glassy, chakraborty2014energy}. This structural complexity, coupled with the significant influence of environmental factors, especially the temperature\cite{chen2000rna}, makes RNA an ideal but difficult test case for Generative AI methods in molecular sciences. Over the past decade, MD simulations and enhanced sampling methods have been increasingly employed to gain insights into RNA’s conformational landscape, though obtaining converged global equilibrium distributions across temperatures remains computationally intensive\cite{zerze2021computational, rahimi2023comparison, boccalini2025exploring, bottaro2016rna}. Here, we show that LaTF may serve as a potential tool for predicting RNA’s temperature dependence in a data-efficient manner.

Specifically, we focus on the ggcGCAAgcc tetraloop (PDB: 1ZIH\cite{jucker1996network}), a well-characterized system suitable for validation against both experimental and computational references\cite{sheehy2010thermodynamic, tan2018rna, zerze2021computational, boccalini2025exploring}. Tetraloops are among the most common and functionally important RNA motifs. Despite their small size, they are highly stable and structurally diverse. In this work we model the temperature-dependent structural ensembles for the GCAA tetraloop (Fig.~\ref{fig:cgaa_tetraloop_300-400}(a)) directly from sequence. Secondary structures are first predicted, followed by stepwise modeling of tertiary structures using bioinformatic approaches. The resulting RNA tertiary structures are solvated in explicit water, and simulations are independently performed at 300 K and 400 K. Iterative reseeding of simulations with generated samples from LaTF trained on-the-fly yields total accumulated simulation times of $\sim189 \mu s$ and $\sim148 \mu s$ at 300 K and 400 K,  respectively (see Methods for more details).

The converged simulation data are used to train the LaTF model, with tilting factor $\tau=4.5$ selected to perfectly accommodate the data from two temperatures (Fig.~\ref{fig:cgaa_tetraloop_300-400}(b) and (c)). Despite employing a large lag time for training ($\Delta t= 200 \textrm{ns}$), significantly longer than that used for the Chignolin protein in Sec. \ref{sec_LaTF_temp_quantify} (we remind the reader that there a lag time of 5ns yielded three states), we identify six metastable states at 300 K (Fig.~\ref{fig:cgaa_tetraloop_300-400}(c)), reflecting RNA's glassy energy landscape. Representatives for each state are obtained by projecting all MD conformers into the IB space, evaluating their likelihood using the NF, and selecting the most probable structure for each state (Fig.~\ref{fig:cgaa_tetraloop_300-400}(a)). The structure from the most populated state aligns well with the NMR structure from the PDB, showing an all-atom RMSD of 0.9\AA. The four-nucleotide loop is observed to exhibit greater flexibility than the helical region, where even conformational fluctuations of single nucleotides give rise to new metastable states. At 300 K, the unfolded state tends to adopt conformation stabilized by base stacking, whereas at 400 K, increased entropy effect drives the unfolded state into a broader, deeper, and more diverse free energy basin (Fig.~\ref{fig:cgaa_tetraloop_300-400}(c)).

The thermodynamic interpolation ability of LaTF allows inference of temperature-dependent FES in the IB space. As shown in Fig.~\ref{fig:cgaa_tetraloop_300-400}(d), increasing temperature gradually broadens the unfolded regions and destabilizes the folded state. Since the LaTF-identified CVs effectively separate conformations by their kinetic differences, we use the IB-space distance as a metric to backmap generated samples to their nearest neighbors in the simulation data. This allows classification of generated conformers into folded or unfolded states and facilitates the derivation of a continuous melting curve for the GCAA tetraloop (Fig.~\ref{fig:cgaa_tetraloop_300-400}(e)). The predicted melting profile aligns well with results from extensive simulated tempering simulations performed using the same force field\cite{tan2018rna}. Although both LaTF and prior computational estimates slightly overpredict the experimental melting temperature\cite{sheehy2010thermodynamic}, the discrepancy likely arises from factors such as force-field limitations, state definitions, or ionic conditions, etc. Meanwhile, the RMSD of generated-and-reconstructed structural ensembles relative to the reference PDB structure across temperatures further supports LaTF’s ability to accurately capture the temperature dependence of RNA structural ensembles (Fig.~\ref{fig:cgaa_tetraloop_300-400}(f)).

Notably, the efficient inference of RNA’s complex temperature-dependent behaviors emerges from combined strengths of representation learning and generative modeling. Due to RNA’s intrinsically rugged energy landscape and the essential role of explicit solvent models in capturing its conformational changes\cite{vangaveti2017advances, sponer2018rna}, prior NF-based generative approaches using implicit solvents fall short in accurately modeling such high-dimensional equilibrium distributions, particularly across temperatures\cite{noe2019boltzmann, dibak2022temperature}.  In contrast, LaTF learns compact yet informative representations of the disordered landscape, optimally leveraging explicit-solvent simulation data to precisely model temperature-dependent FES while enabling interpretation and backmapping to all-atom structures.

\section{Conclusion \& Discussion}

Accurately characterizing the equilibrium distributions of complex molecular systems, as well as their response to environmental factors such as temperature, is essential for uncovering the underlying thermodynamic properties and molecular mechanisms, and projecting these distributions onto physically meaningful low-dimensional representations further facilitates interpretation and understanding. Recent advances in generative AI methods, such as NF models, have shown enormous promise in learning these distributions and providing generative access to configurations. However, in the absence of pre-defined or learnable low-dimensional representations, such methods do not scale well with system size and complexity, limiting their broader applicability.
In this work, we present LaTF, a unified end-to-end framework that integrates SPIB with a NF to enable low-dimensional representation learning, metastable state classification, FES estimation, transition pathway interpolation, and inference of temperature-dependent behaviors within a single workflow. We show that both the joint learning of representations and generative models, and the utilization of an expressive tilted Gaussian prior, are critical to LaTF’s performance. The former enables the model to better identify kinetically meaningful CVs and enhances both the efficiency and accuracy of generations, while the latter creates a structured and interpretable prior space that supports physically realistic interpolations. The incorporation of a temperature-steerable parameter into the tilted prior further broadens LaTF’s applicability, allowing it to predict nontrivial thermodynamic responses from limited data. Applications to diverse systems, including a model potential, protein, multi-body particles, and RNA, demonstrate LaTF’s accuracy and utility.

It is worth noting that there is still room for improvement in the LaTF framework. For instance, all our tested cases empirically suggest that training LaTF with data collected at different temperatures works well. However, there is no theoretical guarantee that the learned representations themselves should remain unchanged with temperature. Although SPIB has demonstrated robustness and can capture essential degrees of freedom even when trained on biased data from enhanced sampling\cite{mehdi2022accelerating, lee2024calculating}, additional steps such as reweighting data or rescaling time steps across temperatures may further enhance its performance\cite{bonati2021deep, yang2018refining}. In our implementation, temperature dependence is introduced through linear interpolation of a steerable parameter in the tilted Gaussian prior, which yields numerically reasonable predictions of temperature-dependent FES. Still, allowing this parameter to be learnable or the input of networks may offer greater flexibility and further improve interpolation quality\cite{herron2024inferring, dibak2022temperature}. 

Beyond the implementations presented in this work, the LaTF framework holds promise for broader applications. For example, it may be extended to study systems exhibiting complex phase behavior or glassy transitions across a range of environmental conditions, and this would likely require more careful design of the prior distribution\cite{van2023boltzmann, lee2025exponentially}. Training LaTF with (reweighted) data from enhanced sampling methods such as replica exchange molecular dynamics trajectories could further improve data efficiency. Moreover, the proposed joint learning framework is not limited to SPIB and NF, alternative dynamical autoencoders such as time-lagged autoencoders\cite{wehmeyer2018time}, variational dynamics encoders\cite{hernandez2018variational}, extended autoencoder\cite{frassek2021extended} and EncoderMap\cite{lemke2019encodermap}, as well as generative models like flow matching\cite{lipman2022flow} or diffusion models\cite{ho2020denoising, song2020denoising}, could be incorporated in a similar architecture. We view LaTF as a foundational approach for unified representation learning and generative modeling, upon which a range of future models and applications can be built and extended depending on specific goals and system types.

\section{Methods}
\textbf{RealNVP Normalizing Flows} NF adopts neural networks to construct learnable, invertible mappings. The specific implementation, including the network architecture and transformation formula, is highly flexible. We employ a sequence of real-valued non-volume preserving (RealNVP) transformations $\{\mathcal{F}_{k}/\mathcal{F}
_{k}^{-1}\}_{k=1}^{n}$ to build the NF\cite{dinh2016density}. RealNVP partitions the encoded IB variables into two channels, $(\mu_1, \mu_2)$, and applies a series of invertible operations (multiplication and addition) to one channel while keeping the other fixed (Fig.~\ref{fig:LaTF_model}). Nonlinearity of the transformation is introduced by using neural networks to parameterize the scaling and translation functions, $S_{\theta}$ and $T_{\theta}$, respectively:

\begin{small}
\begin{gather}\label{realnvp}
    \mu_1 \leftarrow \mathcal{F}_{k}(\mu_1, \mu_2) = \mu_1 \\
    \mu_2 \leftarrow \mathcal{F}_{k}(\mu_1, \mu_2) = \mu_2\odot\exp\big(S_{\theta}(\mu_1)\big)+T_{\theta}(\mu_1) 
\end{gather}
\end{small}

In the subsequent transformation, $\mu_2$ remains fixed while $\mu_1$ is updated. This alternating scheme ensures that the overall Jacobian matrix remains triangular, allowing efficient computation of the determinant. With an analytical prior and tractable Jacobian, the likelihood of any sample in the IB space can be explicitly evaluated. More details of NF and RealNVP are included in the SI.

\textbf{Exponentially tilted Gaussian Distribution} The normalization factors for the tilted Gaussian prior, $Z_{\tau}$, and the temperature-steerable tilted Gaussian prior, $Z_{\tau, T}$, can be expressed analytically as:
\begin{small}
\begin{align}\label{tilted_normalize_factor}
\begin{split}
    &Z_{\tau} = M(\frac{d}{2}, \frac{1}{2}, \frac{1}{2}\tau^{2}) + \tau \sqrt{2} \frac{\Gamma(\frac{d+1}{2})}{\Gamma(\frac{d}{2})}M(\frac{d+1}{2}, \frac{3}{2}, \frac{1}{2}\tau^{2}) \\
    &Z_{\tau, T} = M(\frac{d}{2}, \frac{1}{2}, \frac{T}{2}\tau^{2}) + \tau \sqrt{2T} \frac{\Gamma(\frac{d+1}{2})}{\Gamma(\frac{d}{2})}M(\frac{d+1}{2}, \frac{3}{2}, \frac{T}{2}\tau^{2})
\end{split}
\end{align}
\end{small}
where $M(a, b, z)=\sum_{n=0}^{\infty}\frac{a^{(n)}}{b^{(n)}}\frac{z^{n}}{n!}$ is the Kummer confluent hypergeometric function, and $a^{(n)}=a(a+1)\cdots(a+n-1)$ denotes the rising factorial. Samples from these priors are drawn using Metropolis Monte Carlo sampling. Additional details are provided in the SI.

\textbf{Details of Molecular Dynamics Simulations} 
We adopt the analytical form of the three-hole potential system from Ref.\cite{metzner2006illustration}. Langevin dynamics of a unit-mass particle on this surface is simulated with a time step of $0.001$, a temperature setting of $1/k_{B}$, and a friction coefficient of $0.5 \ $ step$^{-1}$. The simulation runs for $5\times 10^{7}$ steps, recording coordinates every 50 steps, resulting in a trajectory of  $10^{6}$ frames (see SI for more details). 

For the Chignolin protein, simulations are initialized from the NMR structure (PDB: 1UAO) in explicit solvent, followed by energy minimization and multi-step equilibration, with and without positional restraints. The protein is modeled using the OPLS-AA force field \cite{jorgensen1996development, robertson2015improved}, and the TIP3P water model \cite{mark2001structure} is employed. Six independent, long-time unbiased MD simulations are performed for tens of microseconds at six different temperatures, ranging from 340 K to 440 K. The observation of more than twenty reversible folding-unfolding transitions in each trajectory indicates sufficient sampling (simulation details and trajectory lengths are provided in SI).

For the LJ7 system, we simulate seven identical particles in 2D space interacting via the Lennard-Jones potential. Six independent simulations are performed at temperatures ranging from $0.2\epsilon/k_{B}$ to $0.7\epsilon/k_{B}$, using a Langevin thermostat with a friction coefficient of $0.1\sqrt{\epsilon/m\sigma^{2}}$. Each simulation runs for $10^{7}$ steps, with particle coordinates recorded every 100 steps, yielding trajectories of $10^{5}$ snapshots per temperature (see SI for more details).

For the RNA ggcGCAAgcc tetraloop system, we begin with the sequence alone and predict corresponding secondary structures using ViennaRNA\cite{hofacker1994fast, lorenz2011viennarna}, followed by tertiary structure modeling using the FARFAR method in Rosetta\cite{das2008macromolecular, das2010atomic, watkins2020farfar2}. In unbiased simulations, RNA is modeled with the DESRES-AMBER force field\cite{tan2018rna}, solvated in TIP4P-D water\cite{abascal2005general}, and neutralized with 1 M KCl, following the setup from prior work\cite{tan2018rna}. Before production runs, the systems undergo energy minimization and stepwise equilibration with and without positional restraints. Hundreds of microseconds simulations are performed at 300 K and 400 K over a few adaptive iterations. After each iteration, an LaTF model is trained using the accumulated data to identify high-likelihood conformers and transition-region structures for reseeding the next round of simulations. Final trajectories are reweighted using validated MSMs to ensure that the data used for LaTF training are both thermodynamically and kinetically unbiased. Full details on structural modeling, simulation procedures, adaptive sampling, and MSMs construction and validation are provided in the SI.

\textbf{Single-Temperature LaTF Training} The LaTF training procedure follows that of SPIB\cite{wang2021state, wang2024information}, with the key difference being the omission of the on-the-fly VampPrior update, which is replaced by a more expressive NF model (see pseudocode in SI). Training LaTF at a single temperature requires molecular descriptors and an initial set of state labels as input. For the model potential system, the $x$–$y$ coordinates are used as input, and initial labels are generated via K-means clustering. For the Chignolin protein (340 K), we use pairwise distances between carbon-alpha atoms as descriptors and initialize labels through projection onto Time-lagged Independent Components\cite{naritomi2011slow, noe2015kinetic} followed by K-means clustering. During training, LaTF iteratively merges short-lived states to make their lifetimes longer than the lag time $\Delta t$ and relabels input configurations to improve state metastability. We find that a two-step training strategy, i.e., first training SPIB to obtain a converged state assignment, followed by joint training of SPIB and NF, largely reduces training time and effort. This approach is recommended and used in this study; and we confirm that it yields results consistent with end-to-end joint training for all systems presented in this study. 

For both systems, the long simulation trajectory is uniformly divided into five segments, with four used for training and the remaining one reserved for validation. The KL divergence between the generated latent distribution and that encoded from the validation dataset is used to select the optimal tilting factor $\tau$ for the prior. After training the LaTF models, both training and validation data are assigned to metastable states based on the highest decoding probability, allowing construction and evaluation of corresponding MSMs using the GMRQ score\cite{mcgibbon2015variational}. Additional details on feature extraction, label initialization, neural network architecture, and training hyperparameters are provided in the SI.

\textbf{Multi-Temperature LaTF Training} The primary distinction between multi-temperature and single-temperature training lies in the inclusion of an additional input: the temperature-steerable parameter assigned to each configuration. We define this parameter such that the lowest training temperature is set to one, with other temperatures rescaled proportionally. While alternative normalizations may also be viable, the optimal tilting factor $\tau$ in the tilted Gaussian prior should then be adjusted to minimize the KL divergence between the generated and validation-data-encoded latent distributions.

The Chignolin system is featurized using pairwise distances between carbon-alpha atoms, while the RNA tetraloop is represented by $\bm{r}$-vectors\cite{bottaro2014role} and selected pairwise carbon distances. For both systems, initial labels are generated by applying Principal Component Analysis to multi-temperature data, followed by K-means clustering. In the LJ7 system, each configuration is described by sorted coordination numbers, and initial labels are assigned based on K-means clustering over their second and third moments. For all cases, the dataset is uniformly split into five partitions, with four used for training and one for validation. More details on feature construction, label initialization, network architecture, and training hyperparameters are provided in the SI.

\section{Data \& Code Availability}
The simulation data associated with this research is available upon request. The source code for Latent Thermodynamic Flows model and associated documentation is available at \url{https://github.com/tiwarylab/LatentThermoFlows.git}.

\section{Acknowledgements}
This research was entirely supported by the US Department of Energy, Office of Science, Basic Energy Sciences, CPIMS Program, under Award No. DE-SC0021009. The authors thank Dr. Ruiyu Wang for assistance with the Lennard-Jones 7 simulation setup, and Akashnathan Aranganathan for valuable discussions. We thank UMD HPC’s Zaratan and NSF ACCESS (project CHE180027P) for computational resources. P.T. is an investigator at the University of Maryland Institute for Health Computing, which is supported by funding from Montgomery County, Maryland and The University of Maryland Strategic Partnership: MPowering the State, a formal collaboration between the University of Maryland, College Park, and the University of Maryland, Baltimore.

%

\clearpage


\section*{Supplementary material}

# Supplementary Information: Latent Thermodynamic Flows: Unified Representation Learning and Generative Modeling of Temperature-Dependent Behaviors from Limited Data

Yunrui Qiu,[1, 2] Richard John,[1, 3] Lukas Herron,[1, 4, 2] and Pratyush Tiwary [*1, 2, 5]

[1]*Institute for Physical Science and Technology, University of Maryland, College Park, MD, 20742, USA*
[2]*Institute for Health Computing, University of Maryland, Bethesda, MD, 20852, USA*
[3]*Department of Physics, University of Maryland, College Park, MD, 20742, USA*
[4]*Biophysics Program, University of Maryland, College Park, MD, 20742, USA*
[5]*Department of Chemistry and Biochemistry, College Park, MD, 20742, USA*

## Contents

---

[*] Corresponding author:ptiwary@umd.edu

# I. A UNIFIED LOSS FRAMEWORK FOR LATENT THERMODYNAMIC FLOWS

The State Predictive Information Bottleneck (SPIB)[1] has been developed and validated as a state-of-the-art method for the systematic analysis of molecular dynamics (MD) data across a variety of systems[2–4]. Grounded in the Information Bottleneck (IB) principle[5, 6], SPIB simultaneously learns an informative low-dimensional latent representation and identifies long-lived metastable states. Here, we integrate a generative model, specifically a normalizing flow, into the latent space, which can be trained jointly with SPIB. The low-dimensional representation allows the generative model to focus its expressivity on the most meaningful attributes of the data and accelerates sample generation. On the other hand, the generative model enables accurate quantification of the latent distribution, thereby improving the optimization of the latent space. Furthermore, we find that, with an informative prior, the latent generative model exhibits strong generative capability in reproducing free energy landscapes across a broad range of temperatures despite being trained on limited data from only a few temperature conditions. We refer to the whole framework, including both representation learning and generative modeling, as Latent Thermodynamic Flows (LaTF). This section provides a brief overview of the IB principle, reviews the objective function for SPIB, and presents a detailed derivation for the loss function of LaTF.

## A. Information Bottleneck (IB)

The IB principle is designed to extract a low-dimensional representation $\boldsymbol{z}$ from high-dimensional input data $\boldsymbol{X} \in \{\boldsymbol{X}^n\}_{n=1}^N$ that retains the most relevant information to predict a target variable $\boldsymbol{y} \in \{\boldsymbol{y}^n\}_{n=1}^N$. More formally, we presuppose some conditional probability distribution $p_\theta(\boldsymbol{z}|\boldsymbol{X})$, parameterized by $\theta$, which we call an *encoder*. The IB principle consists of maximizing the following objective with respect to the encoder parameters:

$$\underset{\theta}{argmax} \ \mathcal{L}_{IB} = I(\boldsymbol{z}, \boldsymbol{y}; \theta) - \beta I(\boldsymbol{X}, \boldsymbol{z}; \theta) \tag{1}$$

where $I(x, y) = \int dx dy \ p(x, y) \log \frac{p(x, y)}{p(x)p(y)}$ denotes mutual information, which quantifies the shared information between the variables $x$ and $y$. Models maximizing this objective function thus seek to determine a latent representation $\boldsymbol{z}$ that balances being maximally informative when predicting $\boldsymbol{y}$, but relies on the minimum amount of information about $\boldsymbol{X}$, a tradeoff mediated by a hyperparameter $\beta \in [0, +\infty)$.

Due to the high computational cost of directly quantifying mutual information, a variational lower bound is commonly employed as an alternative objective function. To motivate this variational lower bound, we first note that for any probability distributions $q(\boldsymbol{y}|\boldsymbol{z})$ and $r(\boldsymbol{z})$, the following inequality must be satisfied due to the positivity of the Kullback-Leibler divergence and Markovianity conditions on $\boldsymbol{X}$, $\boldsymbol{y}$, and $\boldsymbol{z}$:

$$\mathcal{L}_{IB} \geq \mathcal{L} = \int d\boldsymbol{X} d\boldsymbol{y} d\boldsymbol{z} \Big[ p(\boldsymbol{X}) p(\boldsymbol{y}|\boldsymbol{X}) p_\theta(\boldsymbol{z}|\boldsymbol{X}) \log q(\boldsymbol{y}|\boldsymbol{z}) \Big] - \beta \int d\boldsymbol{X} d\boldsymbol{z} \Big[ p(\boldsymbol{X}) p_\theta(\boldsymbol{z}|\boldsymbol{X}) \log \frac{p(\boldsymbol{z}|\boldsymbol{X})}{r(\boldsymbol{z})} \Big] + \mathcal{H}(\boldsymbol{y}) \tag{2}$$

where the entropy $\mathcal{H}(\boldsymbol{y}) = -\int d\boldsymbol{y} \ p(\boldsymbol{y}) \log p(\boldsymbol{y})$ is a constant and can be omitted during the optimization process. We may now parameterize the *decoder* $q_\theta(\boldsymbol{y}|\boldsymbol{z})$ and *prior* $r(\boldsymbol{z})$ distributions in preparation for maximizing this lower bound on the true objective. Finally, we may eliminate the inaccessible distributions $p(\boldsymbol{X})$ and $p(\boldsymbol{y}|\boldsymbol{X})$ by approximation via the empirical distribution $p(\boldsymbol{X}, \boldsymbol{y}) = \frac{1}{N} \sum_n \delta_{\mathbf{X}^n}(\boldsymbol{X}) \delta_{\boldsymbol{y}^n}(\boldsymbol{y})$. The final variational lower bound to the information bottleneck objective function is as follows:

$$\mathcal{L}_{IB} \geq \mathcal{L} \approx -\frac{1}{N} \sum_{n=1}^N \int d\boldsymbol{z} \Big[ -p_\theta(\boldsymbol{z}|\boldsymbol{X}^n) \log(q_\theta(\boldsymbol{y}^n|\boldsymbol{z})) \Big] - \beta \frac{1}{N} \sum_{n=1}^N \int d\boldsymbol{z} \Big[ p_\theta(\boldsymbol{z}|\boldsymbol{X}^n) \log \frac{p_\theta(\boldsymbol{z}|\boldsymbol{X}^n)}{r(\boldsymbol{z})} \Big] \tag{3}$$

which clearly depends only on the distributions $p_\theta(\boldsymbol{z}|\boldsymbol{X})$, $q_\theta(\boldsymbol{y}|\boldsymbol{z})$ and $r(\boldsymbol{z})$, and empirical data $\{\boldsymbol{X}^n, \boldsymbol{y}^n\}$. Thus, by appropriately choosing analytical forms for these three distributions and optimizing the entire objective with respect to $\theta$, we arrive at a latent variable $\boldsymbol{z}$ satisfying the information bottleneck principle.

## B. State Predictive Information Bottleneck (SPIB)

The time-sequential nature of MD data inspired the extension of the IB framework into SPIB. In line with the IB principle, SPIB aims to extract the most predictive features from molecular structures, enabling accurate inference of their future evolution. Specifically, the inputs to SPIB are high-dimensional structural descriptors representing MD conformations, and the outputs are labels corresponding to the metastable state into which the conformation is expected to transit after a lag time $\Delta t$. As a result, the low-dimensional bottleneck space typically integrates structural descriptors with strong time-lagged correlations and effectively captures the essential slowest dynamical motions. In particular, SPIB employs deep neural networks to parameterize the probability distributions $p_\theta(z|X)$, $q_\theta(y|z)$ and $r_\theta(z)$, and is typically trained on unbiased MD trajectories, with state labels derived from clustering strategies. For instance, given an unbiased trajectory $\{X^{dt}, X^{2dt}, \cdots, X^{Mdt}, \cdots, X^{(M+s)dt}\}$ and the corresponding state labels $\{y^{dt}, y^{2dt}, \cdots, y^{Mdt}, \cdots, y^{(M+s)dt}\}$, where $s \cdot dt = \Delta t$, the objective function of SPIB is expressed as:

$$\underset{\theta}{argmax} \ \mathcal{L} = \frac{1}{M} \sum_{n=1}^{M} \int dz^n p_\theta(z^n|X^n) \Big[ \log q_\theta(y^{n+s}|z^n) - \beta \log \frac{p_\theta(z^n|X^n)}{r(z^n)} \Big]. \tag{4}$$

The first term, quantifying the predictive accuracy of the target state, is evaluated using a decoder neural network $\mathbb{D}$ with softmax outputs, where the conditional probability $q_\theta(y^{n+s}|z^n)$ is computed as $\sum_{i=1}^{k} y_i^{n+s} \log \mathbb{D}_i(z^n)$, with $y$ denoting an $k$-dimensional one-hot state label vector. The second term acts as a regularization term, measuring the discrepancy between the prior distribution and the encoded posterior. Specifically, a Gaussian encoder $\mathbb{E}$ and a modified variational mixture of posteriors prior are employed, i.e., $p_\theta(z^n|X^n) = \mathcal{N}(z^n; \mathbb{E}_1(X^n), \mathbb{E}_2(X^n)I)$, $r(z) = \sum_{i=1}^{k} \omega_i p_\theta(z|X_{rep}^i)$. The encoded posterior distribution is modeled as a Gaussian, parameterized by the transformed input representation, while the prior distribution is a weighted linear combination of posteriors from representative inputs, subject to the constraint that the weights sum to one ($\sum_{i=1}^{k} \omega_i = 1$). In practice, SPIB is trained using an iterative, self-consistent scheme, wherein the number and positions of states are updated dynamically to enhance state metastability. This setup ensures that each configuration is likely to remain within the same state over the specified lag time $\Delta t$.

## C. Latent Thermodynamic Flows (LaTF)

Within the SPIB framework, the latent space distribution is regularized using a mixture of Gaussian components, which has been shown to improve model expressivity and promote a more structured latent representation compared to the conventional isotropic Gaussian prior. However, the assumptions of a Gaussian encoder and a mixed Gaussian prior are strong and often difficult to justify for complex data such as MD trajectories, thereby hindering accurate characterization of the latent distribution. Here, inspired by prior studies[7–10], we integrate a flow-based model into the SPIB framework to establish an invertible transformation between the latent posterior distribution and a tractable, simple prior distribution, thereby enabling more accurate quantification of the latent space distribution. This unified approach enables simultaneous learning of both informative representations and their associated distributions, further facilitating inference across varying thermodynamic conditions.

Specifically, we employ a normalizing flow $\mathcal{F}_\theta$ to construct the posterior distribution by transforming a random variable $u$, which depends on encoded $X$, into the target latent variable $z$, thereby avoiding the need for an explicitly defined analytical form. The model architecture is illustrated in Figure 1 of the main text. Mathematically, the invertible mapping defined by the normalizing flow can be expressed using an integral involving the Dirac delta function:

$$p_\theta(z|X^n) = \int du \ \delta\Big(z - \mathcal{F}_\theta(u)\Big) p_\theta(u|X^n) \tag{5}$$

Substituting the revised expression for the posterior distribution into the SPIB loss function yields:

$$\mathcal{L}_{LaTF} = \frac{1}{M} \sum_{n=1}^{M} \int d\boldsymbol{z}^n \left[ p_\theta(\boldsymbol{z}^n|\boldsymbol{X}^n) \log \frac{q_\theta(\boldsymbol{y}^{n+s}|\boldsymbol{z}^n) \cdot r^\beta(\boldsymbol{z}^n)}{p_\theta^\beta(\boldsymbol{z}^n|\boldsymbol{X}^n)} \right] \tag{6}$$

$$= \frac{1}{M} \sum_{n=1}^{M} \iint d\boldsymbol{z}^n d\boldsymbol{u} \left[ \delta(\boldsymbol{z}^n - \mathcal{F}_\theta(\boldsymbol{u})) p_\theta(\boldsymbol{u}|\boldsymbol{X}^n) \log \frac{q_\theta(\boldsymbol{y}^{n+s}|\boldsymbol{z}^n) \cdot r^\beta(\boldsymbol{z}^n)}{(\int d\boldsymbol{u}' \delta(\boldsymbol{z}^n - \mathcal{F}_\theta(\boldsymbol{u}')) p_\theta(\boldsymbol{u}'|\boldsymbol{X}^n))^\beta} \right] \tag{7}$$

$$= \frac{1}{M} \sum_{n=1}^{M} \int d\boldsymbol{u} \; p_\theta(\boldsymbol{u}|\boldsymbol{X}^n) \log \frac{q_\theta(\boldsymbol{y}^{n+s}|\mathcal{F}_\theta(\boldsymbol{u})) \cdot r^\beta(\mathcal{F}_\theta(\boldsymbol{u}))}{(\int d\boldsymbol{u}' \delta(\mathcal{F}_\theta(\boldsymbol{u}) - \mathcal{F}_\theta(\boldsymbol{u}')) p_\theta(\boldsymbol{u}'|\boldsymbol{X}^n))^\beta} \tag{8}$$

Let $\boldsymbol{h}' = \mathcal{F}_\theta(\boldsymbol{u}')$ and $\boldsymbol{u}'(\boldsymbol{X}^n) = \mathcal{F}_\theta^{-1}(\boldsymbol{h}')$, so that the determinant of the Jacobian matrix associated with this invertible transformation can be expressed as: $\det[\frac{\partial \boldsymbol{h}'}{\partial \boldsymbol{u}'}] = \det[\frac{\partial \mathcal{F}_\theta(\boldsymbol{u}'|\boldsymbol{X}^n)}{\partial \boldsymbol{u}'}]$. One may further simplify the loss function:

$$\mathcal{L}_{LaTF} = \frac{1}{M} \sum_{n=1}^{M} \int d\boldsymbol{u} \; p_\theta(\boldsymbol{u}|\boldsymbol{X}^n) \log \frac{q_\theta(\boldsymbol{y}^{n+s}|\mathcal{F}_\theta(\boldsymbol{u})) \cdot r^\beta(\mathcal{F}_\theta(\boldsymbol{u}))}{(\int d\boldsymbol{h}' \; |\det[\frac{\partial \boldsymbol{u}'}{\partial \boldsymbol{h}'}]| \; \delta(\mathcal{F}_\theta(\boldsymbol{u}) - \boldsymbol{h}') \; p_\theta(\mathcal{F}_\theta^{-1}(\boldsymbol{h}'))^\beta} \tag{9}$$

$$= \frac{1}{M} \sum_{n=1}^{M} \int d\boldsymbol{u} \; p_\theta(\boldsymbol{u}|\boldsymbol{X}^n) \log \frac{q_\theta(\boldsymbol{y}^{n+s}|\mathcal{F}_\theta(\boldsymbol{u})) \cdot r^\beta(\mathcal{F}_\theta(\boldsymbol{u}))}{(|\det[\frac{\partial \boldsymbol{u}}{\partial \boldsymbol{h}}]| \; p_\theta(\mathcal{F}_\theta^{-1}(\mathcal{F}_\theta(\boldsymbol{u})))^\beta} \tag{10}$$

$$= \frac{1}{M} \sum_{n=1}^{M} \int d\boldsymbol{u} \; p_\theta(\boldsymbol{u}|\boldsymbol{X}^n) \log \frac{q_\theta(\boldsymbol{y}^{n+s}|\mathcal{F}_\theta(\boldsymbol{u})) \cdot r^\beta(\mathcal{F}_\theta(\boldsymbol{u}))(|\det[\frac{\partial \mathcal{F}_\theta(\boldsymbol{u})}{\partial \boldsymbol{u}}]|)^\beta}{(p_\theta(\boldsymbol{u}|\boldsymbol{X}^n))^\beta} \tag{11}$$

$$= \frac{1}{M} \sum_{n=1}^{M} \int d\boldsymbol{u} \; p_\theta(\boldsymbol{u}|\boldsymbol{X}^n) \log \left[ q_\theta(\boldsymbol{y}^{n+s}|\mathcal{F}_\theta(\boldsymbol{u})) \cdot r^\beta(\mathcal{F}_\theta(\boldsymbol{u})) \cdot (|\det[\frac{\partial \mathcal{F}_\theta(\boldsymbol{u})}{\partial \boldsymbol{u}}]|)^\beta \right] + const. \tag{12}$$

$$= \frac{1}{M} \sum_{n=1}^{M} \int d\boldsymbol{u} \; p_\theta(\boldsymbol{u}|\boldsymbol{X}^n) \left[ \log q_\theta(\boldsymbol{y}^{n+s}|\mathcal{F}_\theta(\boldsymbol{u})) + \beta \log r(\mathcal{F}_\theta(\boldsymbol{u})) + \beta \log |\det[\frac{\partial \mathcal{F}_\theta(\boldsymbol{u})}{\partial \boldsymbol{u}}]| \right] + const. \tag{13}$$

Following the setup used in SPIB, we define the latent random variable $\boldsymbol{u}$ using a Gaussian encoder $\mathbb{E}$, such that $p_\theta(\boldsymbol{u}|\boldsymbol{X}^n) = \mathcal{N}(\boldsymbol{u}; \mathbb{E}(\boldsymbol{X}^n), \sigma_\theta \boldsymbol{I})$, and apply the normalizing flow transformation as $\mathcal{F}_\theta(\boldsymbol{u}) = \mathcal{F}_\theta(\mathbb{E}(\boldsymbol{X}^n) + \sigma_\theta \cdot \boldsymbol{\xi})$, where $\boldsymbol{\xi}$ is standard Gaussian noise. The normalizing flow is used to refine the encoded distribution to better match the target prior. Furthermore, the distribution $q_\theta(y^{n+s}|\mathcal{F}_\theta(\boldsymbol{u}))$ could be similarly evaluated using a reversible flow and Gaussian decoder, expressed as $\sum_{i=1}^{k} \boldsymbol{y}_i^{n+s} \log \mathbb{D}_i(\mathcal{F}_\theta^{-1}(\mathcal{F}_\theta(\boldsymbol{u})))$, and the prior $r(\mathcal{F}_\theta(\boldsymbol{u}))$ is modeled with a more general form known as the exponentially tilted Gaussian (see details in Sec. II).

Considering a special case, if the invertible transformation is chosen to be a simple linear mapping, i.e., $\mathcal{F}(\boldsymbol{u}|\boldsymbol{X}^n) = \boldsymbol{\mu}(\boldsymbol{X}^n) + \boldsymbol{\sigma}(\boldsymbol{X}^n) \otimes \boldsymbol{u}$, this linear form corresponds to the reparameterization trick, effectively reducing the model to a standard variational autoencoder (VAE). In this case, the determinant of the Jacobian matrix simplifies to:

$$\mathbb{E}_{\boldsymbol{u} \sim p(\boldsymbol{u})}[|\det[\frac{\partial \mathcal{F}(\boldsymbol{u}|\boldsymbol{X}^n)}{\partial \boldsymbol{u}}]|] = -\sum_{i=1}^{dim} \log \sigma_i(\boldsymbol{X}^n) \tag{14}$$

thereby aligning the resulting loss function with that of the canonical VAE[6].

## II. EXPONENTIALLY TILTED GAUSSIAN DISTRIBUTION

To enhance the model's ability to accurately approximate the latent space distribution and make the model more expressive, we adopt a generalized analytical prior known as the exponentially tilted Gaussian[11]. The conventional isotropic Gaussian probability density function $p_0$ in a $\mathbb{R}^{d_z}$ space is given by:

$$r_0(\boldsymbol{z}) = \frac{1}{(2\pi)^{\frac{d_z}{2}} \det(\Sigma)^{\frac{1}{2}}} \cdot \exp(-\frac{1}{2}(\boldsymbol{z}-\boldsymbol{\mu})^T \Sigma^{-1} (\boldsymbol{z}-\boldsymbol{\mu})) \tag{15}$$

where $\boldsymbol{\mu}$ and $\Sigma$ denote the mean vector and covariance matrix, respectively. By applying an exponential tilting to the isotropic Gaussian distribution based on its norm, with tilting parameter $\tau$, the resulting probability density function $r(\boldsymbol{z}, \tau)$ in $\mathbb{R}^{d_z}$ space is defined as:

$$r(\boldsymbol{z}, \tau) = \frac{\exp(\tau||\boldsymbol{z}||)}{Z_\tau} \cdot \frac{\exp(-\frac{1}{2}||\boldsymbol{z}||^2)}{(2\pi)^{\frac{d_z}{2}}} = \frac{\exp(\tau||\boldsymbol{z}||)}{Z_\tau} \cdot r_0(\boldsymbol{z}) \tag{16}$$

$$= \frac{1}{Z_\tau (2\pi)^{\frac{d_z}{2}}} \cdot \exp(-\frac{1}{2}(||\boldsymbol{z}|| - \tau)^2) \cdot \exp(\frac{1}{2}\tau^2) \tag{17}$$

where $Z_\tau = \mathbb{E}_{\boldsymbol{z}\sim\mathcal{N}(\boldsymbol{0},\boldsymbol{I})}[\exp(\tau||\boldsymbol{z}||)]$ is the normalization constant of the distribution. The standard Gaussian distribution emerges as a special case of the tilted Gaussian when the tilting parameter is set to $\tau = 0$. Notably, by completing the square as shown in Equation 17, the tilted Gaussian is revealed to be radially symmetric, attaining its maximum at $|\boldsymbol{z}| = \tau$. The normalization constant for the distribution can be analytically calculated as[11]:

$$Z_\tau = \mathbb{E}_{\boldsymbol{z}\sim\mathcal{N}(\boldsymbol{0},\boldsymbol{I})}[e^{\tau||\boldsymbol{z}||}] = \int_{\mathbb{R}^{d_z}} e^{\tau||\boldsymbol{z}||} \cdot \frac{\exp(-\frac{1}{2}||\boldsymbol{z}||^2)}{(2\pi)^{-\frac{d_z}{2}}} \, d\boldsymbol{z} \tag{18}$$

$$= \sum_{n \; even}^{\infty} \frac{\tau^n d(d+2)\cdots(d+n-2)}{n!} + \sum_{n \; odd}^{\infty} \frac{\tau^n \mu_1 (d+1)(d+3)\cdots(d+n-2)}{n!} \quad (\mu_1 = \sqrt{2}\frac{\Gamma(\frac{d+1}{2})}{\Gamma(\frac{d}{2})}) \tag{19}$$

$$= M(\frac{d}{2}, \frac{1}{2}, \frac{1}{2}\tau^2) + \tau\sqrt{2}\frac{\Gamma(\frac{d+1}{2})}{\Gamma(\frac{d}{2})} M(\frac{d+1}{2}, \frac{3}{2}, \frac{1}{2}\tau^2) \tag{20}$$

where $M(a, b, z) = \sum_{n=0}^{\infty} \frac{a^{(n)}}{b^{(n)}} \frac{z^n}{n!}$ denotes the Kummer confluent hypergeometric function, while $a^{(n)} = a(a+1)\cdots(a+n-1)$ represents the rising factorial.

When training generative models on data collected at different temperatures, the prior distribution should be designed to reflect temperature dependence. A common approach, supported by prior studies, is to use a standard Gaussian prior with variance scaled proportionally to the temperature, which corresponds to the equilibrium distribution of a simple harmonic oscillator at temperature[12–14]. Here, we propose a more generalized and flexible temperature-dependent tilted Gaussian prior, in which the distribution's variance and mode (most probable radius) vary with temperature. This design yields a more informative prior that better captures temperature-dependent behaviors and more accurately distinguishes differences between temperatures. Specifically, the probability density function at temperature $T$, denoted as $r_T(\boldsymbol{z}, \tau)$, is defined as follows:

$$r_T(\boldsymbol{z}, \tau) = \frac{\exp(\tau||\boldsymbol{z}||)}{Z_{\tau,T}} \cdot \frac{\exp(-\frac{1}{2T}||\boldsymbol{z}||^2)}{(2\pi)^{\frac{d_z}{2}}(T)^{\frac{d_z}{2}}} \tag{21}$$

$$= \frac{1}{Z_{\tau,T}} \frac{1}{(2\pi)^{\frac{d_z}{2}}(T)^{\frac{d_z}{2}}} \exp(-\frac{1}{2T}(||\boldsymbol{z}|| - T\tau)^2 \exp(\frac{1}{2}\tau^2 T^2). \tag{22}$$

Completing the square shows that the temperature-dependent tilted Gaussian is radially symmetric, with both the radius of maximum probability and variance varying linearly with temperature. $Z_{\tau,T}$ denotes the temperature-dependent normalization factor, which could be analytically derived:

$$Z_{\tau,T} = \int_{\mathbb{R}^{d_z}} e^{\tau||\boldsymbol{z}||} \ \frac{\exp(-\frac{1}{2T}||\boldsymbol{z}||^2)}{(2\pi)^{d_z/2} \ (T)^{d_z/2}} \ d\boldsymbol{z} \tag{23}$$

$$= \int_0^{+\infty} S_{d_z}(r) \cdot e^{\tau r} \cdot \exp(-\frac{1}{2T}r^2)/(2\pi)^{d_z/2}(T)^{d_z/2} \ dr \tag{24}$$

$$= \int_0^{+\infty} \frac{(2\pi)^{d_z/2}}{\Gamma(\frac{d_z}{2})} r^{n-1} \cdot e^{\tau r} \cdot \exp(-\frac{1}{2T}r^2)/(2\pi)^{d_z/2}(T)^{d_z/2} \ dr \tag{25}$$

$$= M(\frac{d}{2}, \frac{1}{2}, \frac{T}{2}\tau^2) + \tau\sqrt{2T}\frac{\Gamma(\frac{d+1}{2})}{\Gamma(\frac{d}{2})} M(\frac{d+1}{2}, \frac{3}{2}, \frac{T}{2}\tau^2). \tag{26}$$

### III. NORMALIZING FLOWS AND REAL NVP TRANSFORMATIONS

Probabilistic generative models seek to approximate a target probability distribution from a finite number of samples and quickly generate new samples obeying this estimated density. Normalizing flows (NF) constitute one representative class of probabilistic generative models and generally use a series of invertible transformations to bridge a simple prior distribution to an approximation of the target distribution. NF models are trained directly to maximize the likelihood of observing the empirical samples given a set of learnable functions connecting the prior distribution and target distribution. These functions can then rapidly transform samples drawn from the prior distribution to the target distribution.

Specifically, NF models seek to learn an invertible function $\mathcal{F}$ that transforms a sample $\boldsymbol{x}$ from the target distribution $p_{\boldsymbol{x}}$ into a sample $\boldsymbol{z}$ obeying a simple prior distribution $p_{\boldsymbol{z}}$ via $\boldsymbol{z} = \mathcal{F}(\boldsymbol{x})$. Before considering optimization, we note the change in probability due to a change in coordinates $\boldsymbol{z} = \mathcal{F}_{\theta}(\boldsymbol{x})$ ($\theta$ indicates a parameterization of $\mathcal{F}$) is:

$$p_{\boldsymbol{x}}(\boldsymbol{x}; \theta) = p_{\boldsymbol{z}}(\mathcal{F}_{\theta}(\boldsymbol{x})) \left| \det\left( \frac{\partial \mathcal{F}_{\theta}(\boldsymbol{x})}{\partial \boldsymbol{x}^{\top}} \right) \right| \tag{27}$$

where the Jacobian determinant of the transformation shows up on the right-hand side. Taking the logarithm of both sides yields:

$$\log p_{\boldsymbol{x}}(\boldsymbol{x}; \theta) = \log p_{\boldsymbol{z}}(\mathcal{F}_{\theta}(\boldsymbol{x})) + \log \left| \det\left( \frac{\partial \mathcal{F}_{\theta}(\boldsymbol{x})}{\partial \boldsymbol{x}^{\top}} \right) \right|. \tag{28}$$

The above formulation defines a maximum likelihood estimation problem over samples from $p_{\boldsymbol{x}}$. Given a user-defined prior distribution $p_{\boldsymbol{z}}$, and assuming the log-determinant of the Jacobian can be evaluated, one can fully characterize $p_{\boldsymbol{x}}(\boldsymbol{x}; \theta)$ via the change-of-variables formula. This objective can be optimized if the transformation $\mathcal{F}$ is represented as a neural network. New samples can then be generated via $\tilde{\boldsymbol{x}} = \mathcal{F}_{\theta}^{-1}(\tilde{\boldsymbol{z}})$, where $\tilde{\boldsymbol{z}} \sim p_{\boldsymbol{z}}$. If $\mathcal{F}_{\theta}$ is sufficiently expressive and optimization is successful, the generated samples will follow the target distribution $p_{\boldsymbol{x}}$. Meanwhile, the generated probability density can be evaluated using the Jacobian determinant.

The functional form of reversible bridging function $\mathcal{F}_{\theta}$ and the associated neural network architecture varies across different NF implementations. In this work, we adopt the *Real NVP*, or real-valued, non-volume preserving transformations framework [15], which balances simplicity and expressiveness. In Real NVP, $\mathcal{F}_{\theta}$ is constructed as a sequence of discrete, invertible transformations known as *coupling layers*. These layers are designed such that the Jacobian of the overall transformation is triangular, allowing the determinant to be computed efficiently as a simple product of diagonal terms, without cross dependencies. The tradeoff between transformation speed and model expressivity when choosing the functional form of the transformations is a general challenge that various NF implementations approach differently [16, 17].

The Real NVP NF is a composition of $n$ coupling layers:

$$\boldsymbol{z} = \mathcal{F}_{\theta}(\boldsymbol{x}) = (\boldsymbol{f}_{\theta}^{n} \circ \boldsymbol{f}_{\theta}^{n-1} \circ \cdots \circ \boldsymbol{f}_{\theta}^{1})(\boldsymbol{x}) \tag{29}$$

An example of one coupling layer $\boldsymbol{y} = \boldsymbol{f}_{\theta}(\boldsymbol{w})$ is as follows, where $\boldsymbol{w}_{1:D}$ is an input vector and $\boldsymbol{y}_{1:D}$ is an output vector, both with $D$ components (note that the input $\boldsymbol{w}$ will necessarily only correspond to the actual sample $\boldsymbol{x}$ at the first coupling layer $\boldsymbol{f}_{\theta}^{1}$):

$$\boldsymbol{y}_{1:d} = \boldsymbol{w}_{1:d} \tag{30}$$

$$\boldsymbol{y}_{d+1:D} = \boldsymbol{w}_{d+1:D} \odot \exp(\boldsymbol{S}_{\theta}(\boldsymbol{w}_{1:d})) + \boldsymbol{T}_{\theta}(\boldsymbol{w}_{1:d}) \tag{31}$$

where $\odot$ is the element-wise multiplication operation and the multi-valued scaling function $\boldsymbol{S}_{\theta}$ and translation function $\boldsymbol{T}_{\theta}$ are parameterized by neural networks. For layer-wise likelihood maximization, we require the determinant of the following matrix:

$$\frac{\partial \boldsymbol{f}_\theta}{\partial \boldsymbol{w}^\top} = \frac{\partial \boldsymbol{y}}{\partial \boldsymbol{w}^\top} = \begin{bmatrix} \mathbf{I}_{d \times d} & \mathbf{0} \\ \frac{\partial \boldsymbol{y}_{d+1:D}}{\partial \boldsymbol{w}_{1:d}^\top} & \operatorname{diag}(\exp \boldsymbol{S}_\theta(\boldsymbol{w}_{1:d})) \end{bmatrix} \tag{32}$$

These coupling layers alternate so that after each one, only a subset $\{d+1 : D\}$ of the components of the input vector have changed. We can then switch the order in the next layer so that the components with indices $\{d+1 : D\}$ stay unchanged, while $\{1 : d\}$ are transformed. The Jacobians of the transformations are either upper- or lower-triangular, so the determinant is simply the product along the main diagonal. The overall Jacobian determinant reduces to the product of the determinants of the individual layers, allowing one to write the negative log-likelihood of one data sample $\boldsymbol{x}$ under a given set of model parameters as:

$$\mathcal{L} = -\log p_{\boldsymbol{x}}(\boldsymbol{x}; \theta) = -\log p_{\boldsymbol{z}}(\mathcal{F}_\theta(\boldsymbol{x})) - \sum_{k=1}^n \log \left| \exp \sum_i (\boldsymbol{S}_{\theta,i}^k(\boldsymbol{w}_{1:d}^k)) \right| \tag{33}$$

where $\boldsymbol{w}_{1:d}^k$ is the input to $\boldsymbol{S}_\theta^k$ at coupling layer $k$, and $i$ indexes the components of $\boldsymbol{S}_\theta^l$. We can easily extend this single-sample loss function to batches by summing the losses due to the additive property of the log-likelihood. The simple yet powerful Real NVP architecture can be made more expressive by increasing the number of coupling layers or the depth of the neural networks at each layer. While we find Real NVP adequate for our purposes, other attractive architectures, such as GLOW [18] and Neural Spline Flows [19], approach balancing expressivity and evaluation speed differently. GLOW was tested as a candidate for integration with SPIB, but its channel-wise convolutions proved less suitable for the low-dimensional latent data of this study than for its intended image domain.

## IV. DETAILS OF MOLECULAR DYNAMICS SIMULATIONS

### A. Three-hole Potential System

The analytical expression for the 2-dimensional three-hole potential is given by:

$$V(x,y) = 3e^{-x^2-(y-\frac{1}{3})^2} - 3e^{-x^2-(y-\frac{5}{3})^2} - 5e^{-(x-1)^2-y^2} - 5e^{-(x+1)^2-y^2} + 0.2x^4 + 0.2(y-\frac{1}{3})^4 \tag{34}$$

We perform MD simulation of a single particle with mass $m = 1$ on this three-hole potential energy surface. The dynamics are propagated using the Langevin middle integrator[20] with a time step of 0.001. The system temperature is maintained at $1/k_B$ via a Langevin thermostat employing a friction coefficient of 0.5 $step^{-1}$. To confine the particle within the 2-dimensional simulation domain, reflective boundary conditions are applied, restricting the particle to the range $x \in [-2.3, 2.3]$, $y \in [-1.7, 2.6]$. The simulation is performed for a total of $5 \times 10^7$ integration steps, with particle coordinates recorded every 50 steps, resulting in a trajectory consisting of $10^6$ frames.

### B. Chignolin Protein

Chignolin (GYDPETGTWG) is a 10-residue $\beta$-hairpin miniprotein, originally engineered by Honda et al.[21] based on a consensus sequence derived from turn motifs structurally analogous to those found in the GB1 hairpin domain. The NMR-resolved structure of Chignolin (PDB ID: 1UAO)[21] is employed as the initial conformation for our MD simulations. The protein is first solvated in a dodecahedral simulation box containing 2,300 explicit TIP3P water molecules[22] and then neutralized with 100 mM NaCl. The box dimensions are selected to ensure that protein residue is positioned at least 1.3 nm away from the box boundaries, and periodic boundary conditions are applied along all dimensions. The OPLS-AA force field[23, 24] is utilized for the protein, as it has recently been benchmarked to deliver the most reliable performance for this system[25]. In the simulations, electrostatic interactions are truncated at a distance of 1.2 nm, with long-range contributions treated using the Particle Mesh Ewald (PME) method[26]. Van der Waals interactions are similarly computed with a cutoff of 1.2 nm. All simulations are performed using a 2 fs integration time step, with all covalent bonds involving hydrogen atoms constrained using the LINCS algorithm[27].

To evaluate the generative capability of the LaTF model across a range of temperatures, we conduct six independent ultra-long MD simulations at 340 K, 360 K, 380 K, 400 K, 420 K, and 440 K, respectively. Prior to the production runs, the system is first subjected to energy minimization for 5,000 steps using the steepest descent algorithm, followed by a multi-stage equilibration protocol at different target temperatures. The equilibration protocol consists of five sequential phases: (i) a 2 ns $NVT$ equilibration in which all protein atoms are restrained to their energy-minimized positions using a harmonic potential with a force constant of 1000 $kJ/(mol \cdot nm^2)$; (ii) a 2 ns $NPT$ equilibration using Berendsen pressure coupling[28] with the same positional restraints; (iii) a subsequent 2 ns $NPT$ equilibration with a reduced restraint force constant of 100 $kJ/(mol \cdot nm^2)$; (iv) another 2 ns $NPT$ equilibration with a further reduced force constant of 10 $kJ/(mol \cdot nm^2)$; and finally, (v) a 2 ns $NPT$ equilibration using Parrinello–Rahman pressure coupling[29] with all restraints removed. After energy minimization and equilibration, $NVT$ production simulations are performed from the equilibrated configurations at six distinct temperatures. The coordinates of the Chignolin protein are recorded every 0.5 ns, and the trajectory lengths corresponding to each temperature are listed in Table S1. We observe that each trajectory exhibits more than 20 reversible folding–unfolding transitions, indicating that all simulations effectively sample the global energy landscape of the Chignolin protein. All Chignolin protein simulations are performed with GROMACS[30].

Table S1: Summary of molecular dynamics trajectory lengths for Chignolin simulated at different temperatures.

| Temperature | 340K | 360K | 380K | 400K | 420K | 440K |
|---|---|---|---|---|---|---|
| Trajectory Length ($\mu s$) | 35.04 | 35.25 | 34.98 | 17.39 | 35.63 | 35.66 |

### C. Lennard-Jones 7 System

The Lennard-Jones 7 (LJ7) system is a widely used model for studying colloidal rearrangements[31–33]. It comprises seven particles confined to a 2-dimensional space, interacting via the Lennard-Jones potential. The LJ7 system exhibits pronounced temperature-dependent behavior, characterized by a marked disparity between the configurational spaces sampled at low and high temperatures. Therefore, this model may serve as an effective benchmark for evaluating the performance of LaTF, particularly in its ability to accurately capture key temperature-dependent behaviors from limited data. We perform six independent MD simulations of the LJ7 system across a range of temperatures, from $0.2\epsilon/k_B$ to $0.7\epsilon/k_B$, in increments of $0.1\epsilon/k_B$. Specifically, the integration time step is set to $0.005\sqrt{m\sigma^2/\epsilon}$, and a Langevin thermostat with a friction coefficient of $0.1\sqrt{\epsilon/m\sigma^2}$ is employed to maintain constant temperature throughout each simulation. The cutoff distances for force calculations and neighbor list construction are chosen as $2.5\sigma$ and $3.0\sigma$, respectively. For each temperature, we perform a simulation of $10^7$ steps and record the coordinates of all particles every 100 steps, resulting in a trajectory comprising $10^5$ snapshots. All LJ7 simulations are performed with PLUMED[34].

### D. GCAA RNA Tetraloops

The hyperstable GCAA tetraloop (ggcGCAAgcc) has been widely used as a benchmark system for evaluating advances in RNA force fields and sampling methodologies. We first generate an ensemble of diverse structures using RNA structure prediction tools to comprehensively sample its rugged energy landscape. Specifically, the thermodynamics-based secondary structure prediction tool ViennaRNA [35, 36] is employed to predict both the most stable and suboptimal secondary structures of the RNA using sequence information alone, resulting in three different secondary structures. The Fragment Assembly of RNA with Full Atom Refinement (FARFAR) method in Rosetta[37–39] is employed to sample corresponding tertiary structures using Monte Carlo sampling combined with a knowledge-based scoring function, resulting in 3,000 tertiary conformations. Subsequently, all 3,000 structures are characterized using G-vector descriptors [40], projected onto the top four principal components via Principal Component Analysis (PCA), and grouped into 20 clusters using the K-means algorithm. The structures closest to the geometric center of each cluster are selected as the initial seeds for MD simulations.

We perform independent, unbiased MD simulations at 300 K and 400 K, respectively. The simulation system is constructed with a dodecahedral box containing the initial RNA structure, 3,900 water molecules, and is neutralized with 1 M KCl. The box size is chosen to ensure all RNA atoms are positioned at least 1.3 nm from the box boundaries, and periodic boundary conditions are applied across every dimension. To model the GCAA tetraloop, we employ the recently developed classical RNA force field DESRES-AMBER[41], an enhanced extension of the AMBER ff14 force field[42], which has been benchmarked through extensive simulations across various RNA systems[41, 43]. Following the recommended protocol, we use the TIP4P-D water model[44] and apply the CHARMM22 ion parameters[45]. The cutoff for Coulombic interactions and van der Waals interactions is consistently set to 1.2 nm, with long-range electrostatics computed using the PME method[26]. The integration time step is chosen as 2 fs, and all bonds involving hydrogen atoms are constrained using the LINCS algorithm[27]. Following the same equilibration protocol used in the Chignolin protein simulations described above, we perform energy minimization, a 2 ns $NVT$ relaxation with positional restraints, three successive 2 ns $NPT$ relaxations with gradually reduced restraint forces, and a final 2 ns $NPT$ relaxation without restraints. Production $NVT$ simulations are then initiated from the equilibrated configurations, and RNA coordinates are recorded every 1.0 ns.

The interpolation and generative capabilities of the LaTF model offer a powerful tool for reseeding simulations and conducting adaptive sampling. Specifically, we conduct four iterations of adaptive sampling, each comprising 20 independent simulations. Between each iteration, the LaTF model is trained on the accumulated data collected so far at 300 K and 400 K, using an appropriate lag time to enable the identification of 10 metastable states in the latent space. For the next iteration, we select the lowest free energy configuration from each state and midpoints interpolated between each pair of nearest low–free-energy configurations as starting points for new simulations (see details in Sec. V (D)). A total of 80 trajectories at 300 K and 80 trajectories at 400 K were obtained, with average lengths of $\sim 2.37\mu s$ and $\sim 1.85\mu s$, respectively; the distributions of trajectory lengths are shown in Figure S7 and S8.

Given the inherently rugged and complex energy landscape of RNA, it is challenging to directly infer a converged

landscape from parallel short simulations, particularly at lower temperatures. To ensure that the data used to train the LaTF model is thermodynamically and kinetically unbiased, we generate long-timescale trajectories at two temperatures using Markov State Models (MSMs) for training. For each temperature, we first represent tetraloop structures from MD simulations using $r$-vectors and carbon atom pairwise distances (see Sec. V(B)), and project them into a 5-dimensional kinetic space using Time-lagged Independent Component Analysis (TICA)[46] with kinetic mapping [47]. The projected conformations are then clustered into 200 microstates using the K-means algorithm, and MSM is constructed with a 200 ns lag time. While MSM hyperparameters could be further optimized using cross-validation with the generalized matrix Rayleigh quotient (GMRQ)[48] or Variational approach to Markov Processes (VAMP) score[49], we adopt this setup to focus on generating ultra-long unbiased trajectories and validate the models using Implied Timescale (ITS) analysis and the Chapman-Kolmogorov (CK) test[50] (see Figure S7 and S8).

## V. TRAINING OF LATENT THERMODYNAMIC FLOWS

### A. Feature Extraction and Initialization of State Labels

#### Three-hole Potential System

The $(x, y)$ coordinates are provided as input to LaTF, and the initial labels are obtained by applying K-means clustering to the trajectories in the $(x, y)$ space, resulting in 100 clusters.

#### Chignolin Protein

Chignolin protein structures are featurized using pairwise distances between carbon-alpha atoms of residue pairs that are separated by at least two positions along the sequence, resulting in a 28-dimensional feature vector. For the single-temperature trainings at 340 K, initial labels are generated using TICA[46] with kinetic mapping[47] followed by K-means clustering, which has proven effective for SPIB training of protein systems[2]. Specifically, three independent components are selected based on spectral gap analysis[51], and the projected data are partitioned into 100 clusters. For training with data from two temperatures, initial labels are generated using PCA followed by K-means clustering. Feature vectors from both temperatures are concatenated and projected onto the top 10 principal components, ensuring that any remaining component contributes less than 1% to the explained variance ratio. The resulting projected structures are then clustered into 100 clusters.

#### Lennard-Jones 7 System

The 2-dimensional configurations of the seven-particle system are described using coordination numbers. Specifically, for each particle $i$, the coordination number descriptor $c_i$ is defined as:

$$c_i = \sum_{j=1, j \neq i}^{7} s_{ij} \tag{35}$$

where $s_{ij}$ represents the switching function applied to the pairwise contact distance $r_{ij}$ between particles $i$ and $j$, defined as:

$$s_{ij} = \frac{1 - (r_{ij}/r_0)^8}{1 - (r_{ij}/r_0)^{16}} \quad (r_0 = 1.5\sigma) \tag{36}$$

The switching function ensures that the coordination number is a smooth, continuous function with well-defined derivatives. To ensure that the structural embedding is invariant to particle permutations, the concatenated coordination number for each snapshot is sorted before being input into the LaTF model.

Previous studies[31, 33] have shown that the second and third moments of the coordination numbers, $\mu_2^2 = \frac{1}{7} \sum_{i=1}^{7} (c_i - \langle c_i \rangle)^2$ and $\mu_3^3 = \frac{1}{7} \sum_{i=1}^{7} (c_i - \langle c_i \rangle)^3$, serve as informative and meaningful order parameters (OPs) for the LJ7 system. Initial labels for LaTF training are generated using these two empirical OPs. Specifically, trajectories collected at two different temperatures are projected onto $\mu_2^2$ and $\mu_3^3$ and subsequently clustered into 100 clusters via the K-means algorithm.

#### GCAA RNA Tetraloops

The RNA structures are represented using two internal descriptors: $\boldsymbol{r}$-vectors[40] and pairwise distances between ribose $C_4'$ carbon atoms. The $\boldsymbol{r}$-vectors are constructed by concatenating the relative position vectors between the centers of the six-membered rings for all nucleotide pairs. Noting that the distance information captured by $\boldsymbol{r}$-vectors effectively characterizes and distinguishes various interaction patterns such as base stacking and Watson–Crick or non–Watson–Crick base pairing[40], we therefore use only the norms of the $\boldsymbol{r}$-vectors as structural descriptors. As a result, each GCAA RNA structure is represented as a 90-dimensional feature vector, which is further used as input for LaTF training. By concatenating the featurized simulation data from 300 K

and 400 K, we apply PCA to project all RNA configurations onto the top 12 principal components, such that the explained variance of the remaining components is below 1%. The projected data are then grouped into 100 clusters using the K-means algorithm, yielding the initial labels for LaTF training.

## B. Training Procedure

The training procedure of LaTF majorly mirrors that of SPIB[2], except that the on-the-fly update of the variational mixture of posteriors prior is omitted, as these are replaced by a more expressive normalizing flow architecture. LaTF is also trained in a self-consistent and iterative manner, where the number of states and their corresponding labels are dynamically updated to maximize state metastability. The pseudocode for LaTF training is provided in Algorithm 1.

In practice, we observe that when the number of initial states is relatively large, the training process of the whole LaTF model requires a significantly long time to reach loss convergence during the first few refinements. An effective strategy to accelerate the training process is to first pretrain the SPIB part (i.e., the encoder and decoder) using the vanilla SPIB objective and protocol to obtain a reasonably converged set of metastable states. Afterward, the normalizing flow component is incorporated, and all modules are trained jointly with the full LaTF loss function. We note the results obtained from these two training procedures are found to be highly consistent across all systems examined in this study.

---

**Algorithm 1** Latent Thermodynamic Flows

---

**Input:** MD simulation trajectories at one or multiple temperatures $\{\mathbf{X}^i_{(T_i)}\}_{i=1}^L$, corresponding set of initial state labels $\{\mathbf{y}^i_{(T_i)}\}_{i=1}^L$, encoder $\mathbb{E}_\theta$ and decoder $\mathbb{D}_\theta$ networks, normalizing flow model $\mathcal{F}_\theta$ with a specified number of coupling layers, lag time $\Delta t$, convergence threshold $thd$, convergence patience $n_{\text{patience}}$ for stopping criteria, number of refinement iterations $n_{\text{refinements}}$, tilted Gaussian prior parameter $\tau$

1: **Preprocessing:** create training dataset such that each sample consists of $\{\mathbf{X}^i_{(T_i)}, \mathbf{y}^{i+\Delta t}_{(T_i)}, T_i\}$
2: Set $\text{loss}_0, \text{loss}_1 \leftarrow 0$
3: **for** $iter$ from 1 to $n_{\text{refinements}}$ **do**
4:     **repeat**
5:         Set $\text{loss}_0 = \text{loss}_1$
6:         Sample a batch of training data $\{\mathbf{X}^i_{(T_i)}\}$, $\{\mathbf{y}^{i+\Delta t}_{(T_i)}\}$ and $\{T_i\}$
7:         $\text{loss}_1 \leftarrow$ calculate the objective function $\mathcal{L}_{LaTF}$, which includes the prediction accuracy from $\{\mathbf{X}^i_{(T_i)}\}$ to future state $\{\mathbf{y}^{i+\Delta t}_{(T_i)}\}$, and a regularization term enforcing the mapping of latent representations to $\{T_i, \tau\}$ temperature-steerable tilted Gaussian priors via the normalizing flow
8:         Update the parameters $\theta$ of neural networks including the encoder, decoder and normalizing flow
9:     **until** the condition $|\text{loss}_0 - \text{loss}_1| < thd$ holds for $n_{\text{patience}}$ consecutive updates
10:     Update the state labels $\{\mathbf{y}^{(T_i)}_i\}$ via $\hat{\mathbf{y}}^i_{(T_i)} = \underset{i}{\text{argmax}} \; \mathbb{D}_i(\mathbb{E}_\theta(\mathbf{X}^i_{(T_i)}); \Delta t, \theta)$
11: **end for**

---

## C. Network Architecture and Training Hyperparameters

To optimize training efficiency, we recommend a two-stage training protocol for the LaTF model. First, pretrain the vanilla SPIB using its original loss function to obtain a reasonably converged metastable state classification. Then, jointly train the encoder, decoder, and normalizing flow using the LaTF objective. All hyperparameters used in the two-stage training protocol are listed in Table S2 and Table S3.

Two critical hyperparameters in LaTF are the dimension of the latent space and the tilted Gaussian parameter $\tau$. As demonstrated in prior studies[2], unlike dimensionality reduction methods based on the VAMP theory [49], which extract orthogonal slow modes and thus can only separate a limited number of metastable states per component, SPIB can accommodate a substantially larger number of metastable states even within a 2-dimensional latent space[2]. Accordingly, all results presented in this manuscript are consistently obtained using a 2-dimensional latent space for LaTF. Meanwhile, the choice of the tilted prior parameter $\tau$ is intended to achieve a clear separation while maintaining appropriate overlap between data collected at different temperatures. So the optimal value of

Table S2: Neural network architectures and training hyperparameters used for pretraining the State Predictive Information Bottleneck (SPIB) across the different systems presented in this study.

| System | SPIB Parameters | | | | | | | | |
|---|---|---|---|---|---|---|---|---|---|
| | Learning Rate | $\beta$ Weight | Encoder Neurons | Decoder Neurons | Lagtime $\Delta t$ | Batch Size | Convergence Threshold | Convergence Patience | Refinement Number |
| Three-Hole Potential ($k_B T = 1.0$) | $10^{-3}$ | $10^{-4}$ | 16 | 16 | 1500 steps | 512 | $10^{-3}$ | 3 | 10 |
| Chignolin Protein (340K) | $10^{-3}$ | $3 \times 10^{-4}$ | 32 | 32 | $30 ns$ | 512 | $10^{-3}$ | 3 | 15 |
| Chignolin Protein (340K & 440K) | $10^{-3}$ | $10^{-4}$ | 32 | 32 | $5 ns$ | 512 | $10^{-3}$ | 5 | 15 |
| Chignolin Protein ($1\mu s$ traj, 340K & 440K) | $10^{-2}$ | $5 \times 10^{-4}$ | 32 | 32 | $5 ns$ | 256 | $10^{-3}$ | 10 | 15 |
| Chignolin Protein (380K & 440K) | $10^{-3}$ | $10^{-4}$ | 32 | 32 | $5 ns$ | 512 | $10^{-3}$ | 5 | 15 |
| Lennard-Jones 7 ($k_B T = 0.2\epsilon$ & $0.5\epsilon$) | $2 \times 10^{-3}$ | $10^{-4}$ | 16 | 16 | 100 steps | 512 | $10^{-3}$ | 5 | 15 |
| Lennard-Jones 7 ($k_B T = 0.2\epsilon$ & $0.7\epsilon$) | $2 \times 10^{-3}$ | $10^{-4}$ | 16 | 16 | 100 steps | 512 | $10^{-3}$ | 5 | 15 |
| CGAA RNA (300K & 400K) | $2 \times 10^{-3}$ | $5 \times 10^{-5}$ | 32 | 32 | $200 ns$ | 512 | 0.001 | 5 | 15 |

Table S3: Neural network architectures and training hyperparameters used for training the Latent Thermodynamic Flows (LaTF) across the different systems presented in this study.

| System | Latent Thermodynamic Flows Parameters | | | | | | | | |
|---|---|---|---|---|---|---|---|---|---|
| | Learning Rate | $\beta$ Weight | Coupling Layers | $\tau$ Value | Flow Neurons | Batch Size | Convergence Threshold | Convergence Patience | Refinement Number |
| Three-Hole Potential ($k_B T = 1.0$) | $10^{-3}$ | $10^{-4}$ | 10 | 3 | 16 | 512 | 0.1 | 200 | 1 |
| Chignolin Protein (340K) | $10^{-3}$ | $3 \times 10^{-4}$ | 35 | 3 | 16 | 512 | 0.1 | 200 | 1 |
| Chignolin Protein (340K & 440K) | $10^{-3}$ | $10^{-4}$ | 35 | 2.5 | 16 | 512 | 0.1 | 150 | 0 |
| Chignolin Protein ($1\mu s$ traj, 340K & 440K) | $10^{-2}$ | $5 \times 10^{-4}$ | 35 | 3 | 16 | 256 | 0.1 | 150 | 0 |
| Chignolin Protein (380K & 440K) | $10^{-3}$ | $10^{-4}$ | 35 | 3.5 | 16 | 512 | 0.1 | 150 | 0 |
| Lennard-Jones 7 ($k_B T = 0.2\epsilon$ & $0.5\epsilon$) | $2 \times 10^{-3}$ | $10^{-4}$ | 20 | 2 | 16 | 512 | 0.1 | 150 | 0 |
| Lennard-Jones 7 ($k_B T = 0.2\epsilon$ & $0.7\epsilon$) | $2 \times 10^{-3}$ | $10^{-4}$ | 20 | 2 | 16 | 512 | 0.1 | 150 | 0 |
| CGAA RNA (300K & 400K) | $2 \times 10^{-3}$ | $5 \times 10^{-5}$ | 50 | 4.5 | 16 | 4096 | 0.1 | 150 | 0 |

$\tau$ is selected by minimizing the Kullback–Leibler (KL) divergence between the generated data and the reference data at the training temperatures. In particular, LaTF is trained using a cross-validation scheme, where different segments of the data are used for training and validation, i.e., the entire dataset is divided into five folds, with four used for training and one reserved for validation. The KL divergence is computed only on the validation sets. The value of $\tau$ that results in the average lowest generation KL divergence across all training temperatures is selected for subsequent use.

## D. Interpolation of Transition Pathways

To interpolate transition pathways using LaTF, we first need to identify source and sink configurations. The normalizing flow in LaTF analytically characterizes the latent distribution, making it well-suited to locate configurations with the lowest free energy. Simultaneously, the decoder assigns any given latent samples to metastable states. Together, they enable the identification of minimum free energy configurations within each metastable state, which serve as the source and sink for pathway interpolation. After flowing the source and sink configurations to the prior space, we employ two commonly used strategies for pathway interpolation: spherical linear interpolation[52] is applied to the LaTF with a standard Gaussian prior, while linear interpolation is used for the LaTF with a tilted Gaussian prior.

Specifically, spherical linear interpolation[52] connects the source $\boldsymbol{z}^{(0)}$ and the sink $\boldsymbol{z}^{(1)}$ via:

$$\boldsymbol{z}^{(\alpha)} = \frac{\sin((1-\alpha)\theta)}{\sin(\theta)}\boldsymbol{z}^{(0)} + \frac{\sin(\alpha\theta)}{\sin(\theta)}\boldsymbol{z}^{(1)} \tag{37}$$

where $\theta = \arccos(\frac{(\boldsymbol{z}^{(0)})^T\boldsymbol{z}^{(1)}}{||\boldsymbol{z}^{(0)}||\ ||\boldsymbol{z}^{(1)}||})$ denotes the angle between source and sink, and $\alpha$ is chosen to uniformly partition the interval $(0,1)$. For the tilted Gaussian prior, due to its well-structured form, we directly employ linear interpolation between the central angles and radii of the source and sink configurations. All interpolated samples are subsequently passed through the inverse normalizing flow and mapped to the latent space.

In the case study of the Chignolin protein, we benchmark the interpolated transition pathways against the kinetic pathways identified via Transition Path Theory[53, 54]. Specifically, the conformations from the 340 K MD trajectory are encoded into the 2-dimensional latent space using the trained LaTF model and subsequently grouped into 200 clusters via the K-centers algorithm. A Markov State Model is constructed with a lag time of 20 ns, selected based on implied timescale analysis and validated using the Chapman–Kolmogorov test[50]. The net flux between each pair of metastable states is computed, and the Dijkstra algorithm is applied to identify all possible transition pathways. The top five flux-carrying pathways, which collectively account for 22% of the total flux, are visualized in the main text.

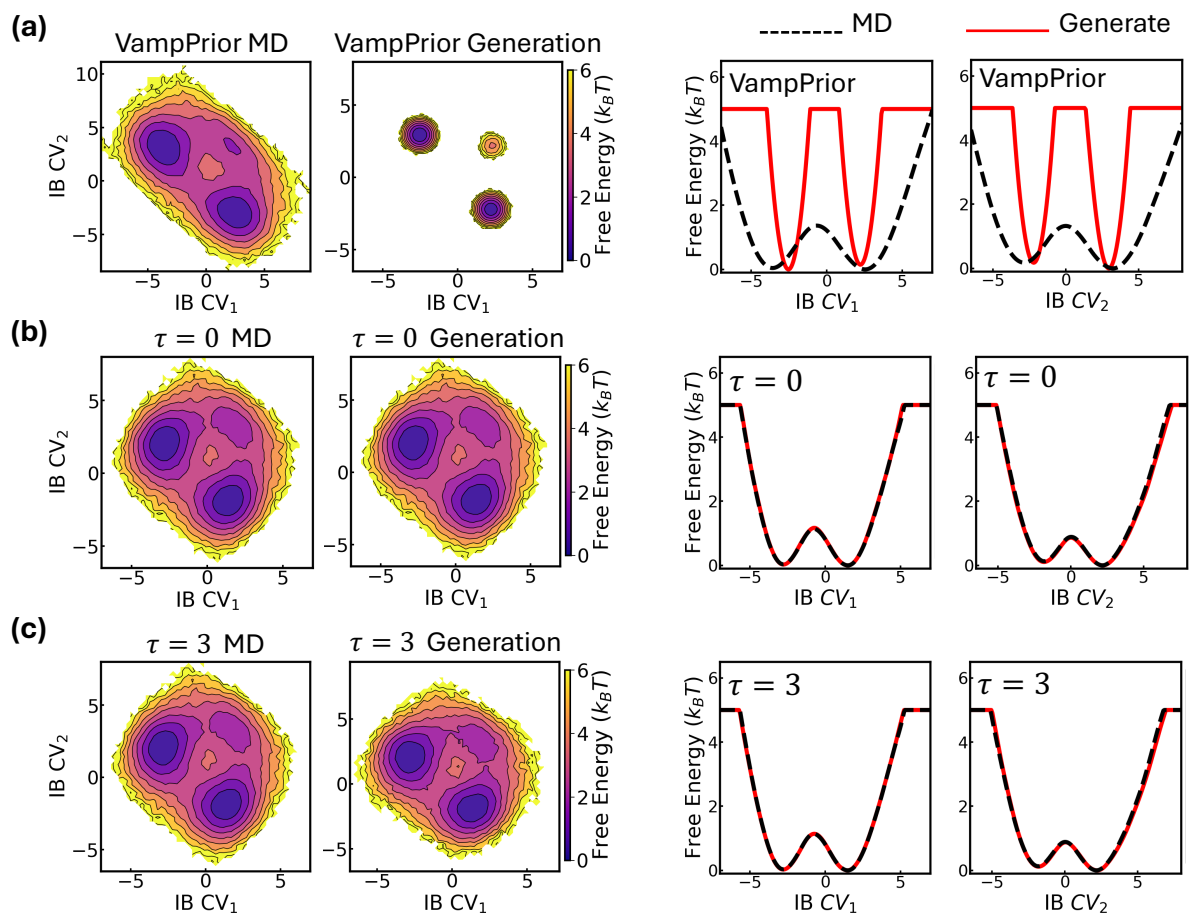

**Figure S1: Comparison of generative performance between VampPrior and Latent Thermodynamic Flows (LaTF) on the 2-dimensional three-hole potential system**. Visualization of the free energy landscapes in the State Predictive Information Bottleneck (SPIB) latent space, estimated directly from encoded MD data or from samples generated by different generative models: (a) vanilla SPIB model using VampPrior; (b) LaTF model with a standard Gaussian prior; and (c) LaTF model with a tilted Gaussian prior, where the tilted parameter is set to $\tau = 3$.

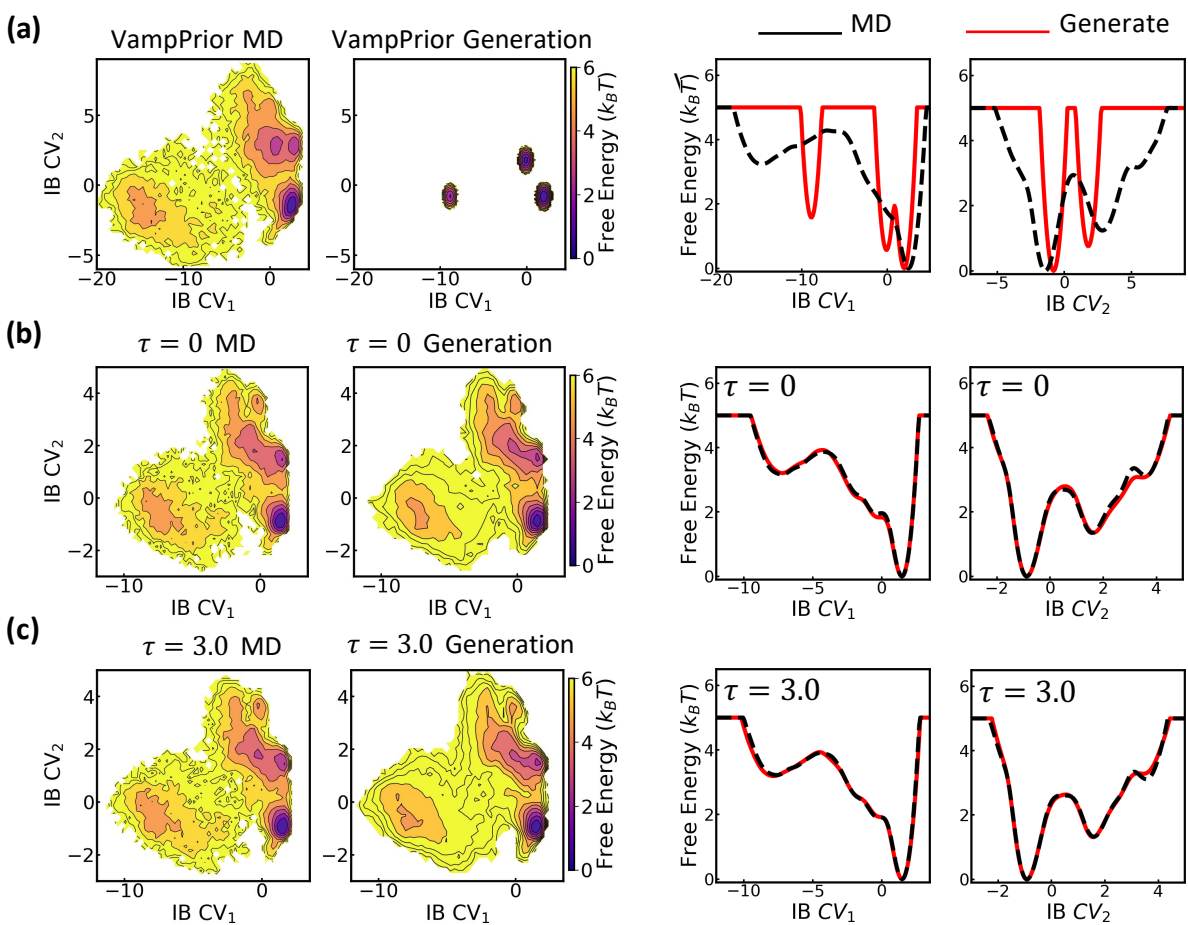

**Figure S2: Comparison of generative performance between VampPrior and Latent Thermodynamic Flows (LaTF) on the Chignolin protein folding system.** Vanilla State Predictive Information Bottleneck (SPIB) and Latent Thermodynamic Flows (LaTF) models with varying tilted prior parameters are trained using MD data of Chignolin at 340 K. Free energy landscapes in the SPIB-encoded latent space, estimated either directly from MD trajectories or from samples generated by LaTF, are visualized for comparison: (a) vanilla SPIB model using VampPrior; (b) LaTF model with a standard Gaussian prior; and (c) LaTF model with a tilted Gaussian prior, where the tilted parameter is set to $\tau = 3.0$.

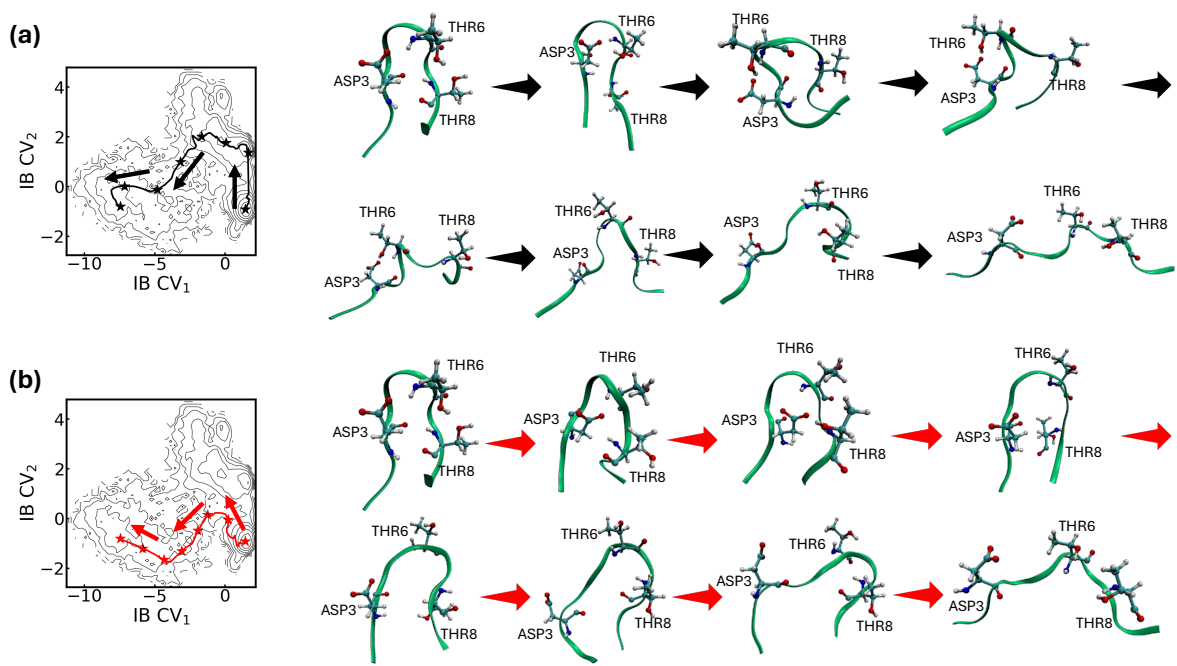

**Figure S3: Two transition pathways interpolated by the Latent Thermodynamic Flows (LaTF) model for the Chignolin protein, along with key intermediate conformations identified along each pathway.** The LaTF model is trained on MD data at 340 K with a tilted prior parameter of $\tau = 3$. The most probable conformations in the latent space from the folded and unfolded states are selected as the source and sink, respectively, with their probabilities quantified using the Jacobian determinant of the normalizing flow transformation. Linear interpolation is performed between the source and sink in the tilted Gaussian prior space, and the interpolated points are mapped back to the latent space. For each pathway, six generated intermediate conformations are selected and reconstructed to all-atom structures using nearest neighbors in the latent space. Three key residues that serve as markers distinguishing the two folding pathways are highlighted. The two interpolated pathways reveal distinct folding mechanisms of the Chignolin protein.

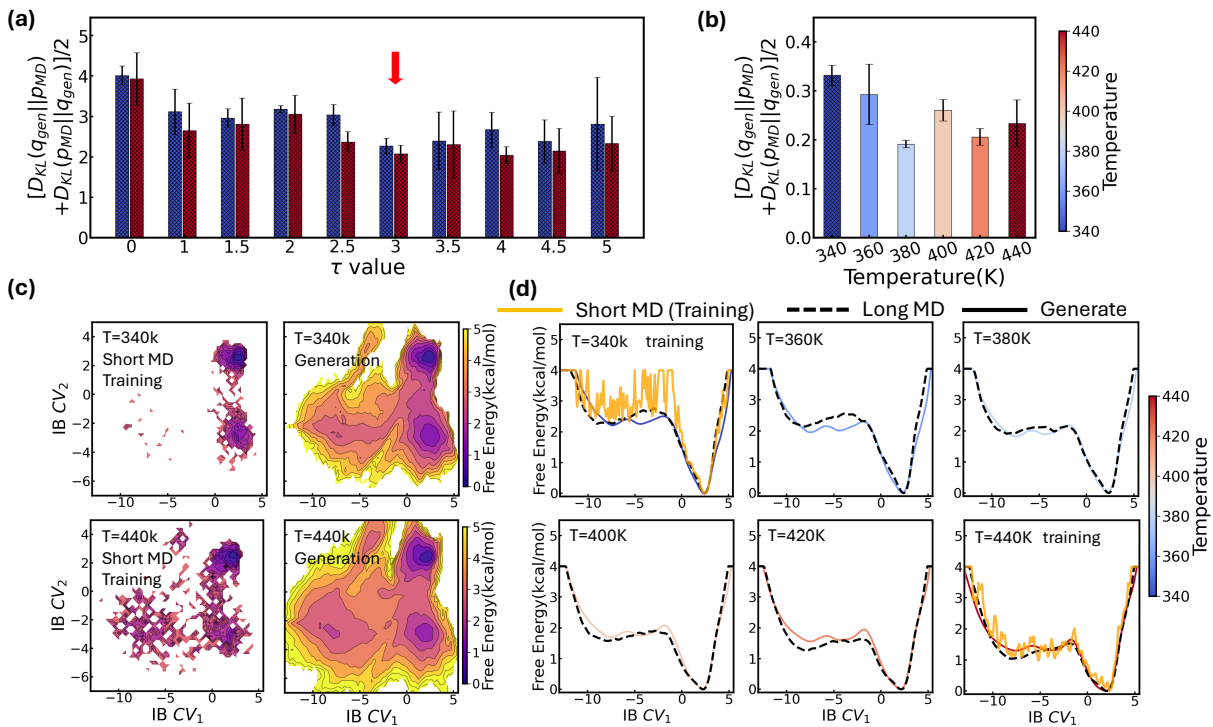

**Figure S4: Benchmarking the generative performance of Latent Thermodynamic Flows (LaTF) trained with highly limited data.** The LaTF model is trained using the first $1\mu s$ segment of Chignolin protein MD trajectory collected at 340 K and 440 K. (a) Symmetrized Kullback–Leibler (KL) divergence between the latent-space distributions of encoded $1\mu s$ MD trajectories and LaTF-generated samples with different prior tilted parameters at the two training temperatures, 340 K and 440 K. The optimal tilted parameter is determined to be $\tau = 3$ (indicated by the red arrow), which results in the lowest KL divergence across the training temperatures. (b) Symmetrized KL divergence between the latent-space distributions generated by LaTF and the reference distributions encoded from full ultra-long MD trajectories across various temperatures. All uncertainties are estimated by training the LaTF model ten times with different random initialization seeds. (c) Free energy landscapes in the latent space estimated from $1\mu s$ training MD trajectories (first column) and from latent samples generated by LaTF models trained on limited data (second column). (d) Comparison of free energy landscapes obtained from $1\mu s$ short MD trajectories (orange solid), ultra-long MD trajectories (black dashed), and LaTF-generated samples (colorful solid).

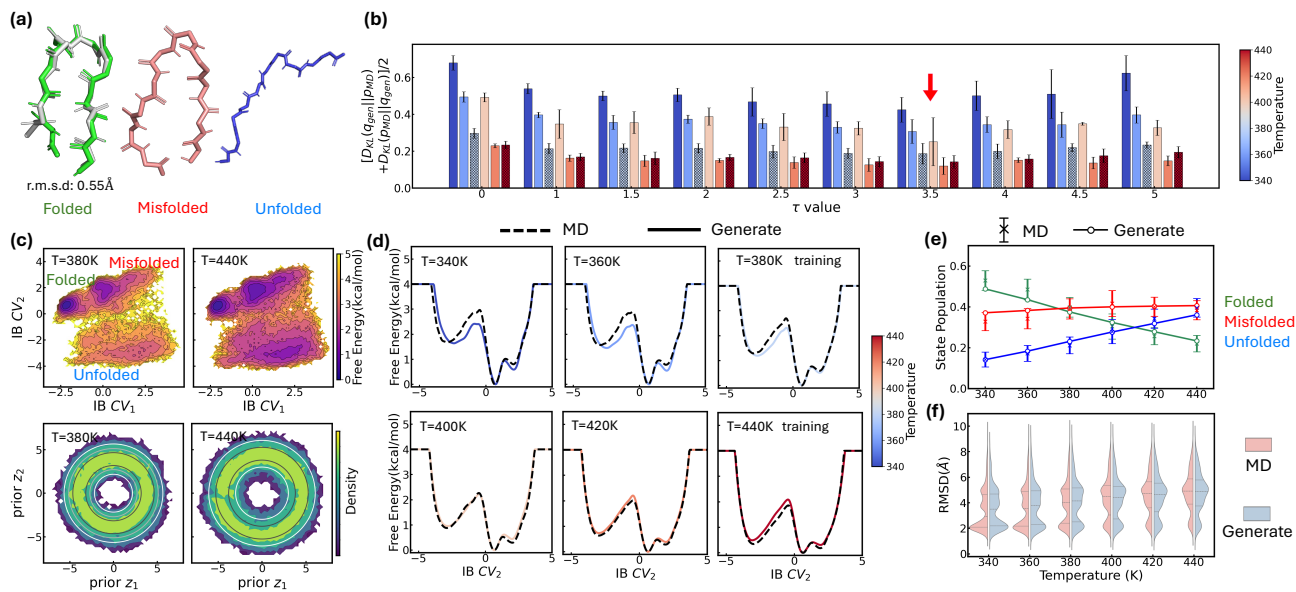

**Figure S5: Benchmarking the generative capability of Latent Thermodynamic Flows (LaTF) in low-temperature regimes when trained on data from comparatively higher temperatures for Chignolin protein.** The LaTF model is trained using data collected at 380 K and 440 K, and its generative performance is evaluated across other temperatures. The MD trajectory at each training temperature is uniformly divided into five segments, with four segments used for training and one for validation. (a) Backbone structures of the highest-probability conformations scored by LaTF ($\tau = 3.5$) for the Chignolin protein, corresponding to different states classified by the State Predictive Information Bottleneck (SPIB). The highest-scoring folded structure (green) is overlaid with the NMR structure (grey), and the backbone RMSD is computed. (b) Symmetrized Kullback–Leibler (KL) divergence between the LaTF-generated distribution and the encoded MD distribution in the SPIB latent space across different temperatures. Training temperatures are indicated by shaded colors. KL divergences are computed using only the validation set at training temperatures and the full dataset at other generative temperatures. The error bars are quantified accordingly using the cross-validation results. The optimal tilted parameter, $\tau = 3.5$, is selected based on KL divergence evaluated on the validation dataset at the training temperatures and is subsequently used to generate all other results (highlighted by the red arrow). (c) Free energy landscapes estimated from the encoded MD data in latent space are shown for 380 K and 440 K, with the corresponding states indicated (top row). The encoded MD data are then transformed forward to the prior distribution, and the resulting distributions are visualized (bottom row). Contour lines derived from the analytical prior are overlaid to assess the performance of the normalizing flow in mapping between the distributions. (d) Comparison between the generative energy landscapes produced by LaTF (solid lines) and the reference obtained from ultra-long MD simulations (dashed lines) across six temperatures. (e) Populations of the three states predicted by LaTF models trained at 380 K and 440 K across a range of temperatures. Error bars indicate state populations estimated from Markov State Models (MSMs) constructed using ultra-long MD trajectories at each temperature, with state assignments based on the trained LaTF. Uncertainties are computed using a Bayesian approach[55]. (f) Comparison of RMSD distributions relative to the NMR structure for MD data and LaTF-generated samples. LaTF samples are reconstructed into all-atom structures using their nearest neighbors in the latent space, which are encoded from the training MD data. Distribution quartiles are marked with dashed lines.

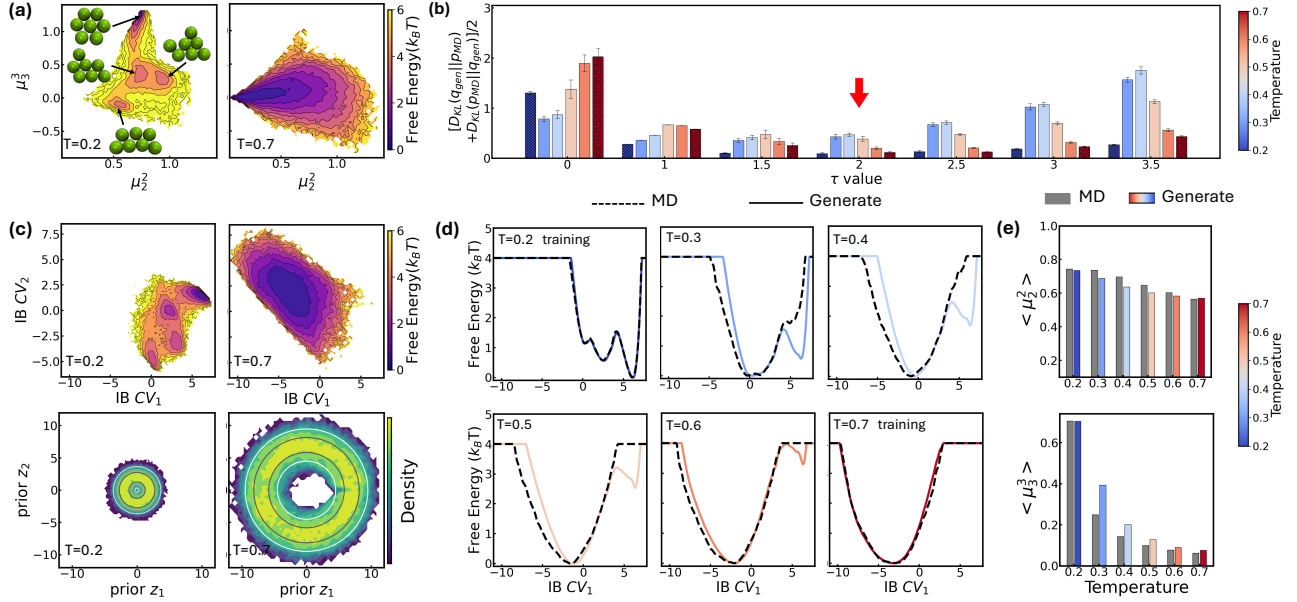

**Figure S6: Benchmarking the generative performance of Latent Thermodynamic Flows (LaTF) trained on datasets acquired at two widely separated temperatures for the Lennard-Jones 7 (LJ7) system.** The LaTF model is trained on data collected at $0.2\epsilon/k_B$ and $0.7\epsilon/k_B$, and used to generate samples at four intermediate temperatures. Long trajectories collected at $0.2\epsilon/k_B$ and $0.7\epsilon/k_B$ are evenly partitioned into five segments, with four segments used for training and the remaining one used for validation. (a) Free energy landscapes projected onto two physical order parameters, i.e., the second and third moments of coordination numbers ($\mu_2^2$ and $\mu_3^3$)[31], estimated from long MD simulations at the two training temperatures. For $0.2\epsilon/k_B$, representative structures corresponding to the highest probability conformations within each metastable state, as scored by LaTF, are also visualized. (b) Symmetrized Kullback–Leibler (KL) divergence between the latent distributions generated by LaTF and the ground-truth distributions encoded directly from MD data. For the two training temperatures, KL divergences are computed only with respect to the validation datasets and highlighted with shading. For all other temperatures, KL divergences are evaluated against the full MD datasets. The optimal prior tilting parameter is identified as $\tau = 2$, as it yields the lowest KL divergence at the training temperatures (highlighted by red arrow). (c) Free energy landscapes projected onto the LaTF latent space using full MD data for the two training temperatures (top row), and corresponding distributions of forward-flowed encoded MD data in the target prior space (bottom row). Contour lines represent the analytical prior and are used to assess LaTF's ability to learn an effective and invertible transformation between the latent and prior distributions. (d) Comparison between the generative free energy landscape (solid line) and the reference free energy landscape obtained from long MD trajectories (dashed line) across six different temperatures. (e) Comparison of the generative (colored bars) and reference (gray bars) means of the second and third moments of coordination numbers across different temperatures. Generative samples are reconstructed into full-coordinate structures by replacing each with its nearest neighbor in the latent space.

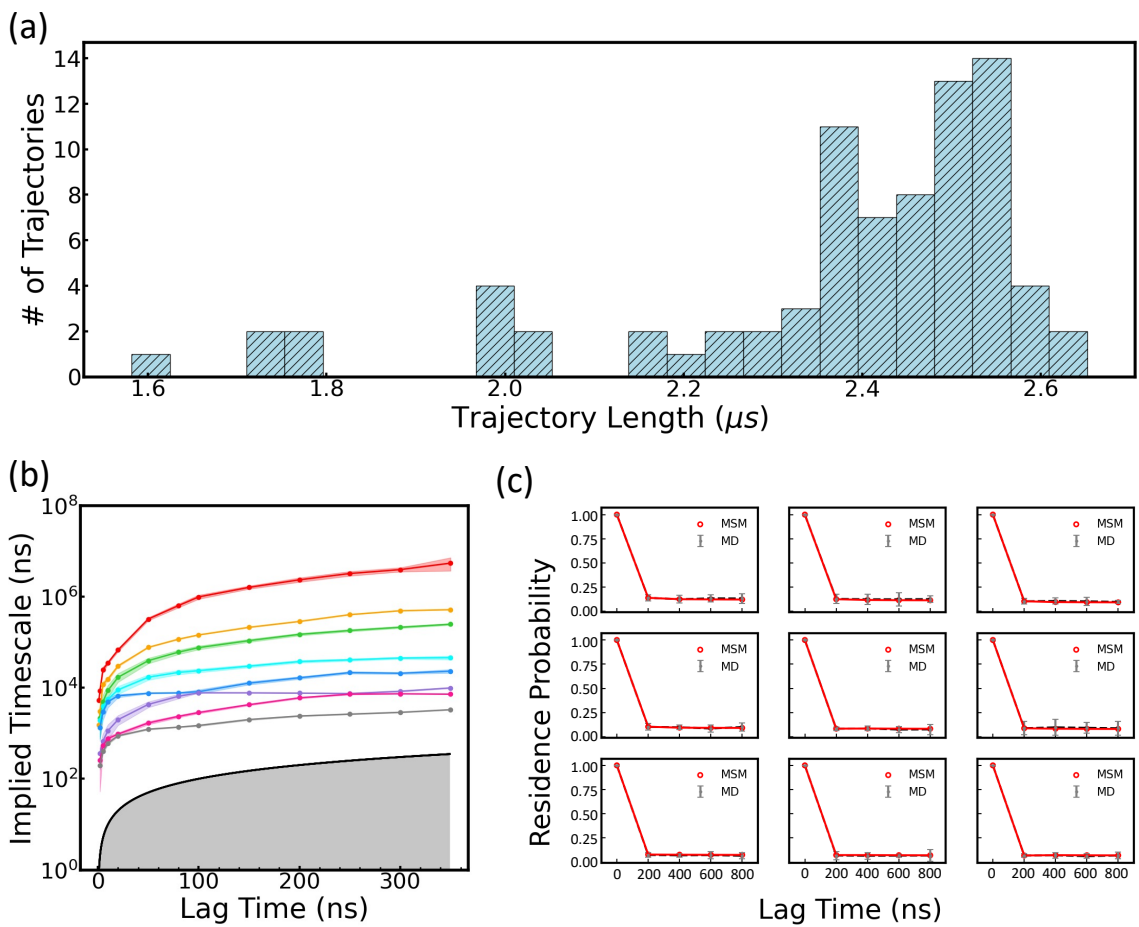

**Figure S7: Construction and validation of the 200-microstate Markov State Model (MSM) using Implied Timescale (ITS) analysis and the Chapman-Kolmogorov (CK) test for GCAA tetraloop RNA at 300 K.** (a) Distribution of simulated MD trajectory lengths for the GCAA tetraloop at 300 K. A total of 80 trajectories were run in parallel, yielding an aggregate simulation time of $\sim 189\mu s$ and an average trajectory length of $\sim 2.37\mu s$. (b) ITS calculated using models constructed with varying lag times. The ITS curves converge beyond a lag time of 200 ns, indicating that models exhibit Markovian behavior with lag times greater than 200 ns. (c) CK test of the MSM constructed with a 200 ns lag time for the nine most populated microstates. The agreement between residence probabilities derived from raw MD data and those predicted by the MSM supports the model's accuracy in capturing long-timescale dynamics. Uncertainties in the ITS and CK test error bars are estimated by bootstrapping the trajectory ensemble 50 times with replacement.

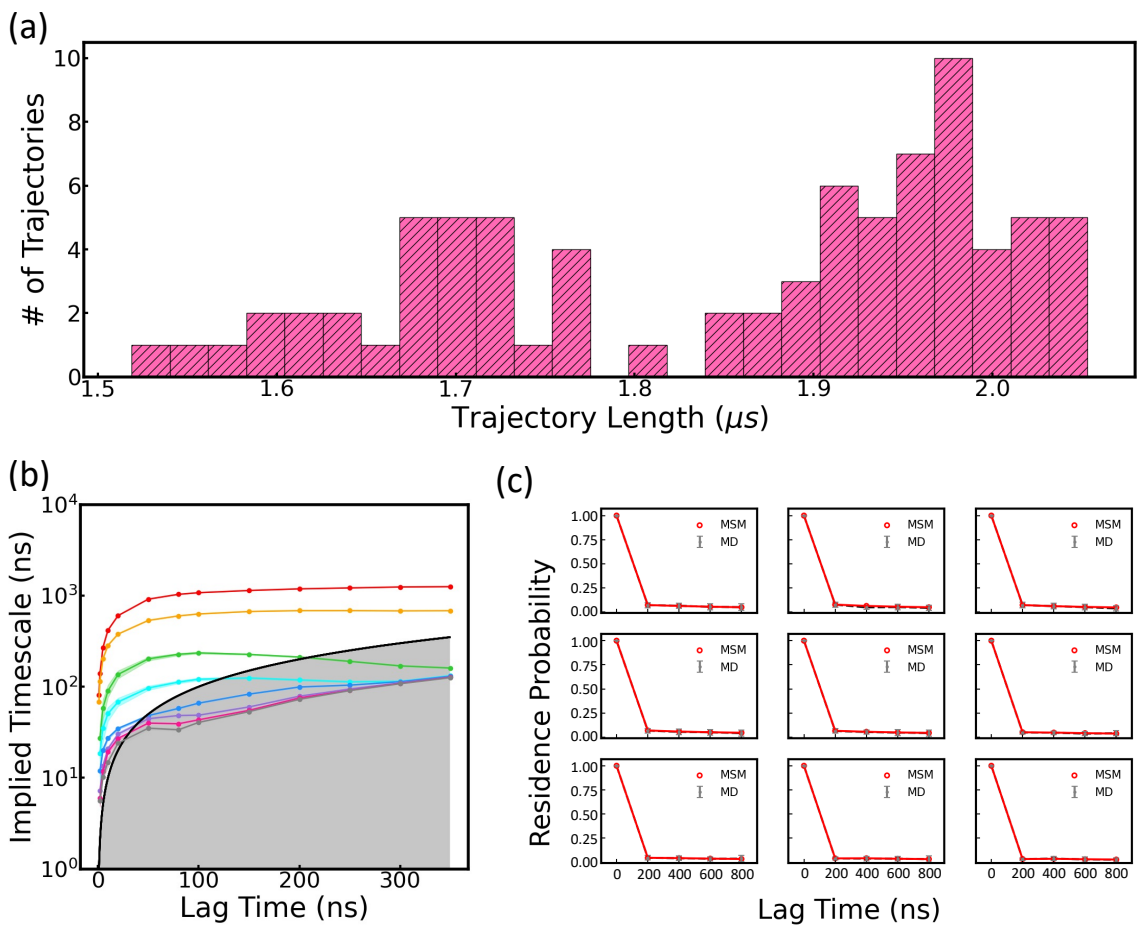

**Figure S8: Construction and validation of the 200-microstate Markov State Model (MSM) using Implied Timescale (ITS) analysis and the Chapman-Kolmogorov (CK) test for GCAA tetraloop RNA at 400 K.** (a) Distribution of simulated MD trajectory lengths for the GCAA tetraloop at 400 K. A total of 80 trajectories were run in parallel, yielding an aggregate simulation time of $\sim 148\mu s$ and an average trajectory length of $\sim 1.85\mu s$. (b) ITS calculated using models constructed with varying lag times. The ITS curves converge beyond a lag time of 200 ns, indicating that models exhibit Markovian behavior with lag times greater than 200 ns. (c) CK test of the MSM constructed with a 200 ns lag time for the nine most populated microstates. The agreement between residence probabilities derived from raw MD data and those predicted by the MSM supports the model's accuracy in capturing long-timescale dynamics. Uncertainties in the ITS and CK test error bars are estimated by bootstrapping the trajectory ensemble 50 times with replacement.



\end{document}